\documentclass[10pt]{article}
\usepackage[T1]{fontenc}
\usepackage{iftex}
\ifPDFTeX\usepackage[utf8]{inputenc}\fi
\usepackage[margin=0.85in]{geometry}
\usepackage{amsmath,amssymb,booktabs,graphicx,array,tabularx}
\usepackage{lmodern}
\usepackage{microtype}
\usepackage{needspace}
\usepackage[font=small,labelfont=bf,labelsep=period,justification=justified,
  singlelinecheck=false]{caption}
\usepackage{subcaption}
\renewcommand{\arraystretch}{1.12}

\usepackage[authoryear,round]{natbib}
\usepackage[hidelinks]{hyperref}
\hypersetup{pdftitle={Synthetic Deformable Tubular Benchmark: A mechanically annotated dataset for evaluating deformable 3D reconstruction},
pdfauthor={Íñigo Cirauqui-Viloria}}
\newcolumntype{Y}{>{\raggedright\arraybackslash}X}
\newcommand{\manuscripttablestyle}{\small\setlength{\tabcolsep}{4pt}\renewcommand{\arraystretch}{1.12}}

\newcommand{\AuthorAffiliation}{Universidad de Zaragoza, Pedro Cerbuna 12, 50009 Zaragoza, Spain}
\newcommand{\CorrespondenceEmail}{icirauqui@gmail.com}
\newcommand{\CorrespondenceContact}{\href{mailto:\CorrespondenceEmail}{\CorrespondenceEmail}}
\newcommand{\AuthorORCID}{0000-0003-1773-1937}
\newcommand{\AuthorORCIDContact}{\href{https://orcid.org/\AuthorORCID}{\AuthorORCID}}
\newcommand{\ManuscriptTitle}{Synthetic Deformable Tubular Benchmark: A mechanically annotated dataset for evaluating deformable 3D reconstruction}
\newcommand{\DatasetVersion}{1.0.0}
\newcommand{\DatasetDOI}{https://doi.org/10.57967/hf/10850}

\newcommand{\LumenMorphVersion}{1.0.2}
\newcommand{\LumenMorphURL}{https://github.com/icirauqui/LumenMorph/tree/v1.0.2}
\newcommand{\ContinuumFEMURL}{https://github.com/icirauqui/continuum-fem}
\newcommand{\ExperimentalCOLMAPURL}{https://github.com/icirauqui/colmap_for_colonoscopy_2/tree/3.7-colonoscopy.1}
\newcommand{\ExperimentalCOLMAPRevision}{feb480d63d5c2025426c162089475492d76fef41}
\newcommand{\OfficialCOLMAPVersion}{4.2.0}
\newcommand{\OfficialCOLMAPForkURL}{https://github.com/icirauqui/colmap/tree/4.2.0-ordering-fixes.1}
\newcommand{\ReproductionPackageURL}{https://doi.org/10.57967/hf/10851}
\newcommand{\PublicationField}[2]{\begingroup\edef\PublicationValue{#1}%
  \ifx\PublicationValue\empty\textbf{[INSERT #2]}%
  \else\expandafter\url\expandafter{\PublicationValue}\fi\endgroup}

\newcommand{\FundingStatement}{This research did not receive any specific grant from funding agencies in the public, commercial, or not-for-profit sectors.}
\newcommand{\CompetingInterestsStatement}{The author declares no competing financial interests or personal relationships that could have appeared to influence the work reported in this paper.}
\newcommand{\AuthorContributionsStatement}{Íñigo Cirauqui-Viloria: Conceptualization, Methodology, Software, Validation, Formal analysis, Data curation, Visualization, Writing---original draft, Writing---review and editing.}
\newcommand{\EthicsStatement}{No human participants, patient data, animals or clinical interventions were used; C3VD is a public static phantom dataset.}
\newcommand{\AIDeclarationStatement}{During the preparation of this work, the author used OpenAI Codex to assist manuscript revision and implementation of analysis, validation and plotting scripts. The client and underlying model versions were not recorded. Numerical results and data visualizations were obtained by executing the documented computational procedures on retained data. The author reviewed and edited the content and takes full responsibility for the content of the article.}

\title{\ManuscriptTitle}
\author{Íñigo Cirauqui-Viloria$^{a}$\\
\small $^{a}$\AuthorAffiliation\\
\small Correspondence: \CorrespondenceContact\\
\small ORCID: \AuthorORCIDContact}
\date{}
\begin{document}
\maketitle
\begin{abstract}
Evaluating deformable 3D reconstruction in tubular environments requires
ground truth that distinguishes camera motion, surface deformation, material
correspondence and image appearance. Existing endoscopic datasets provide
complementary annotations; controlled evaluation additionally benefits from
explicit links between observations, material identities and mechanical states.
We introduce the Synthetic Deformable Tubular Benchmark, generated with
LumenMorph, an open-source toolkit for creating, annotating and evaluating
deformable tubular scenes. The release contains 5,580 RGB images. The baseline
crosses three surface-motion conditions with moving and stationary cameras in
six families; matched appearance variants cover only moving-camera
finite-element actions in four families. Known camera poses, surface geometry, depth,
deformation fields, persistent material correspondences and discrete
finite-element records are accessible through documented public and evaluation
interfaces. Separately implemented geometric and mechanical checks assess
annotation consistency with the generating models. Structure-from-motion,
optical-flow and supplied-geometry displacement-interpolation studies demonstrate
the resource's utility. They distinguish registration coverage from geometric
accuracy, expose camera- and surface-motion effects, and test mechanical priors.
Persistent triangle identities and material weights let users follow the same
surface location across frames, query its three-dimensional trajectory and
evaluate predictions against known motion. Public readers, explicit evaluation
contracts and the configurable generator support reuse and extension beyond
the released acquisitions. Mechanical interpolation illustrates both benefits
and mismatch of a prior, rather than a general ranking of estimators.
The fixed benchmark provides a reusable reference under explicit synthetic
assumptions. Data integrity,
fixed-environment repeatability and retained-prediction score replay have
distinct verification scopes; anatomical and clinical realism are not claimed.

\end{abstract}

{\raggedright\noindent\textbf{Keywords:} Deformable 3D reconstruction; Synthetic dataset; Ground truth; Finite element method; \mbox{Endoscopic imaging}\par}

\section{Introduction}
Deformable reconstruction in colonoscopy must distinguish camera motion from
surface motion under limited visibility. A rigid reconstruction accumulated
over time can absorb deformation into biased camera poses or geometry.
Surface similarity alone may also conceal correspondence errors: a nearby
surface point need not be the same material point---a fixed location on the
wall followed as it moves. Evaluation therefore
requires linked observations, material identities, geometry and visibility,
together with explicit assumptions about the motion-generating model.
Known camera poses alone do not resolve the inverse deformation problem.

\begin{figure}[tbp]
\centering\includegraphics[width=\linewidth]{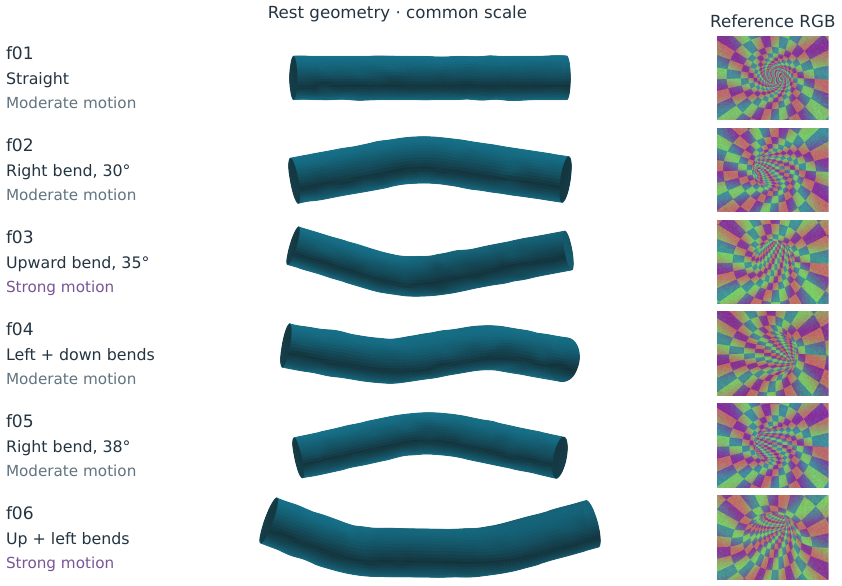}
\caption{Six controlled scene families: one straight tube (f01), single bends
to the right or upward (f02, f03, f05), and two successive bends (f04, f06).
Geometry views share a common scale; the right column shows each family's
reference image with its seeded wall texture. f03 and f06 use twice the
displacement-basis amplitudes of the other families. Table~\ref{tab:population}
specifies bend angles and paired acquisitions.}
\label{fig:atlas}
\end{figure}

We present the Synthetic Deformable Tubular Benchmark, a controlled synthetic
resource \citep{cirauqui2026tubular}. It was generated with \texttt{LumenMorph},
an openly available generator and evaluation toolkit \citep{lumenmorph2026}.
Figure~\ref{fig:atlas} introduces the six scene families:
one straight and five bent tubes, each observed under paired motion and camera
conditions. A matched appearance extension changes lighting, reflectance and
exposure on four of these same families. The resource links material
trajectories, discrete finite-element (FE) records, image annotations and
estimated reconstructions. This connection allows a user to ask not only
whether a surface looks correct, but whether the same material points move
correctly and whether missing surface support explains an apparently low
error. Data, code and reproduction resources are identified in
Sections~\ref{sec:data-availability}--\ref{sec:code-availability}.

The methodological contribution is a benchmark with an explicit, accessible
and checked connection between image observations, camera states, persistent
material identities, deformed geometry and discrete mechanical records. This
integration lets researchers evaluate reconstruction accuracy, material-motion
consistency and surface support under paired acquisition conditions, without
introducing a new reconstruction algorithm. The novelty is the queryable
connection, rather than any one annotation or the use of simulation alone.
The reusable contribution has two parts: the fixed, versioned benchmark,
with linked annotations and separately implemented consistency checks; and
LumenMorph, the open-source generator and reader/evaluation toolkit for
creating and studying additional tubular configurations. Reconstruction,
optical-flow and supplied-geometry mechanical studies illustrate uses of
these resources. They distinguish camera accuracy, sparse geometry, surface
support and registration coverage, while a static phantom control examines
performance on physical imagery under its declared acquisition contract.
The engineered scene families provide controlled reconstruction diagnosis,
not a sampled clinical population.

\subsection{Related work}
Existing resources address complementary evaluation needs. SimCol3D provides
a synthetic depth and pose benchmark with separate anatomies for
training and evaluation~\citep{rau2024simcol3d}, while RealSynCol expands
CT-derived geometry and image realism~\citep{lena2026realsyncol}. C3VD and
C3VDv2 provide phantom images acquired with clinical colonoscopes and
registered dense labels~\citep{bobrow2023c3vd,golhar2026c3vdv2}.
DynamicColon, the dataset introduced alongside the ColonSplat reconstruction
method by \citet{smolak2026colonsplat}, provides
per-time-step point clouds to evaluate whole-scene motion, beyond appearance
along the observed camera path. C3VD-DEFCOL combines parameterized peristalsis and
centerline motion with depth-conditioned image translation, reporting 110
videos from 11 colon geometries and dense dynamic annotations
\citep{luk2026c3vddefcol}. VR-Caps and SimIntestine
extend simulation to capsule
endoscopy~\citep{incetan2021vrcaps,singh2025simintestine}, and SimuScope
incorporates deformable tissues in surgical
simulation~\citep{martyniak2025simuscope}. Dense dynamic annotations and
mechanical simulation are therefore established capabilities. The present
work focuses on their auditable connection to persistent material identity,
controlled acquisitions and reconstruction-pipeline evaluation.

Prior colonoscopy reconstruction methods also motivate this evaluation
setting. Hong et al. used tubular structure to reconstruct virtual colons
from video~\citep{hong2014virtualcolon}; Ruano et al. trained shape-from-shading
depth estimation with synthetic images~\citep{ruano2024sfs}; and Liu et al.
combined synthetic pretraining with intrinsics-guided learning for
scale-consistent monocular reconstruction~\citep{liu2026scale}. Such methods
benefit from benchmarks in which image formation, correspondence and motion
can be examined separately, while clinical transfer is evaluated using
physical data.

Our reconstruction studies use COLMAP's incremental structure-from-motion
(SfM) pipeline~\citep{schonberger2016colmap} through two separately built
implementations: a derivative of COLMAP-for-colonoscopy
\citep{subiascolmap} and a derivative of official release 4.2.0. The upstream
colonoscopy project supplies the starting implementation; the experimental
derivative includes developments described in a subsequent master's
thesis~\citep{cirauqui2022thesis}.
Section~\ref{sec:pipeline-audit} defines the two pipelines explicitly, and
Section~\ref{sec:code-availability} identifies the tested source.
The optical-flow study compares Farneb\"ack and dense inverse search
(DIS)~\citep{farneback2003polynomial,kroeger2016dis} with Recurrent All-Pairs
Field Transforms (RAFT)~\citep{teed2020raft}, using the implementation and
pretrained weights provided by \texttt{torchvision}~\citep{torchvisionraft}.
These are complementary usage studies, not new reconstruction or flow
algorithms.

The motivation is to provide controlled acquisition and deformation conditions
for evaluating reconstruction. The distinguishing feature is the explicit,
queryable connection between image observations, persistent surface material
identities, geometric states and discrete mechanical records. A persistent
triangle and its material weights identify the same surface location across
frames; separate public and evaluator interfaces make ground-truth access
explicit. Discrete loads, responses and energy terms expose the assumptions
behind FE-derived motion. These links allow errors to be examined against
known camera and surface motion while controlling appearance variations.

The controls complement the broader anatomy and appearance of other resources;
we do not demonstrate superior realism or downstream reconstruction performance
against those datasets. The matched appearance extension preserves geometry
and motion rather than adding independent anatomies. LumenMorph also lets
researchers create additional tubular configurations beyond the fixed release
(Section~\ref{sec:extending-lumenmorph}). Table~\ref{tab:related} summarizes
the distinctions from related resources.

\begin{table}[tbp]
\centering\manuscripttablestyle
\begin{tabularx}{\linewidth}{@{}>{\raggedright\arraybackslash}p{0.19\linewidth}YY@{}}
\toprule Resource & Established capability & Relevant distinction \\
\midrule
SimCol3D; RealSynCol & Synthetic colon depth/pose; broader anatomy or appearance & Rich geometric annotations precede this resource. \\
C3VD; C3VDv2 & Real imaging of phantoms; dense registered static labels & C3VDv2 deformation videos lack dense per-step labels. \\
DynamicColon & Deforming geometry with per-step point clouds and depth & Cage-based motion; three reported scenes. \\
C3VD-DEFCOL & Parameterized dynamic geometry and depth-conditioned appearance & Geometric motion; reported biomechanical and appearance-alignment limits. \\
VR-Caps; SimIntestine & Capsule simulation with varied acquisition or peristalsis & Different acquisition setting; complete mechanical export schemas not verified here. \\
SimuScope & Surgical simulation with deformable tissues and instruments & Adjacent anatomy/task; mechanics is established simulation practice. \\
This resource & Pixel-to-triangle/material weights link RGB and cameras to surface trajectories and discrete FE records & Queryable cross-time identity and generating-model checks; appearance flow has a narrower contract. \\
\bottomrule
\end{tabularx}
\caption{Comparison based on primary publications and author release
descriptions. It
is not an independent numerical validation of competing archives.
C3VD-DEFCOL's release repository lists the dataset as forthcoming~\citep{c3vddefcolrelease};
reported capability and downloadable availability are distinct. An
unverified export is not evidence that it does not exist. The distinction claimed
here is an explicit, accessible and checked connection between annotations,
not priority for depth, displacement or FE simulation individually. The six
engineered families do not establish clinical realism.}
\label{tab:related}
\end{table}

\Needspace{9\baselineskip}
\section{Materials and Methods}
The resource has two parts: a baseline bank of 42 image streams and a matched
appearance extension of 20 image streams, for 5,580 color (RGB) images in total.
Every stream shows a camera inside a textured hollow tube. Baseline streams
vary wall motion and camera motion under a checker-texture raster renderer;
extension streams show the same moving-camera FE actions with path-traced
lighting, surface reflectance and exposure variants.
A \emph{family} fixes a seeded tube, texture, camera paths and motion bases;
a \emph{condition} specifies how that tube moves; a \emph{stream} is one
90-frame acquisition. A \emph{package} groups one family's motion condition
with its two action-camera modes and a link to the shared reference scan.
Table~\ref{tab:population} maps every package to its family before those codes
are used below.

\begin{figure}[tbp]
\centering\includegraphics[width=\linewidth]{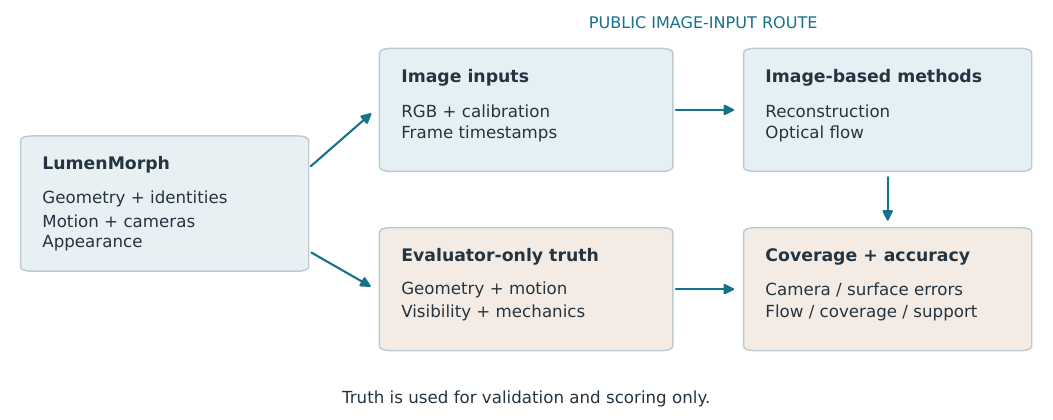}
\par\medskip
\includegraphics[width=\linewidth]{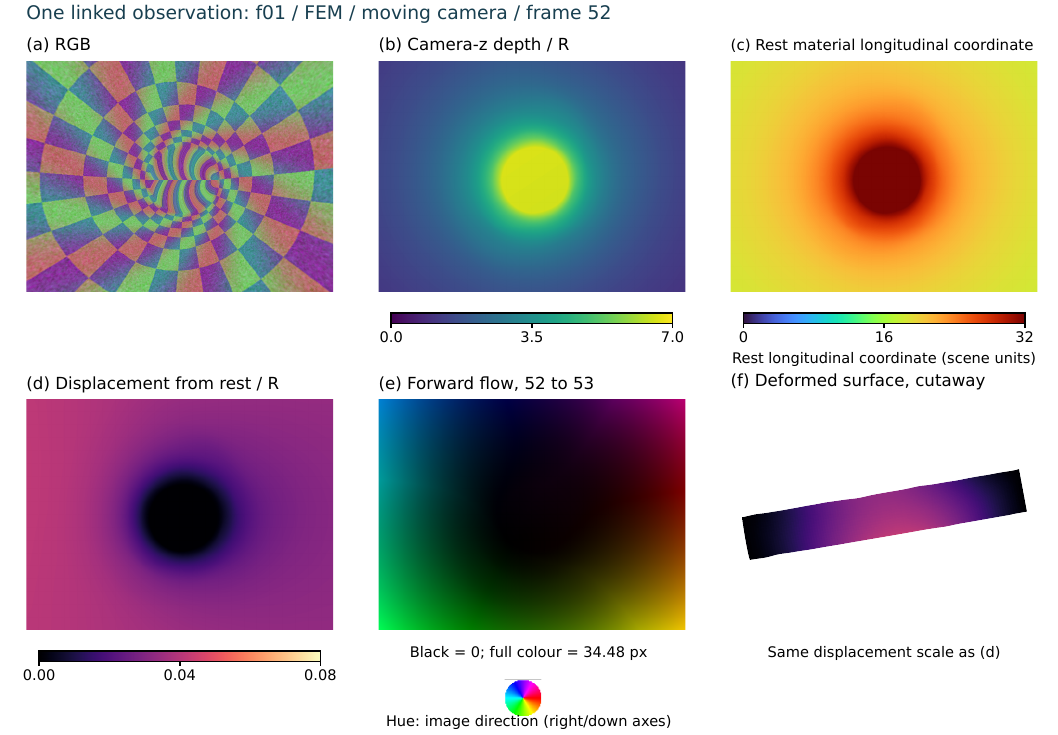}
\caption{Information flow through LumenMorph and the usage studies. Generation
produces both public image inputs and evaluator-only truth. Image-based
methods receive the upper route; their predictions are compared with the
lower route during scoring. Independently implemented validation also checks
truth against the generating geometry and mechanics. Registration, accuracy
and surface support are reported separately, so a missing or incomplete
output cannot masquerade as a low error. Below, one baseline f01 FEM moving-camera
frame (52) links RGB, camera-z depth, persistent material position, displacement
from rest, next-frame flow (52 to 53) and deformed surface geometry. The material
colour is rest longitudinal position, not a new material label; triangle IDs and
barycentrics carry the actual identity. Depth and displacement are normalized
by the reference radius $R$. Flow colour encodes direction and magnitude in
pixels; its key gives the display scale. Background is grey in truth maps.
The surface view cuts away the near wall for display; all annotations use the
complete surface. Codex (version not recorded) assisted the diagram and plotting code.}
\label{fig:overview}
\end{figure}

Figure~\ref{fig:overview} summarizes generation, annotation, validation and
evaluation. LumenMorph defines geometry, material identities and FE states; rendering produces RGB images and dense geometric annotations for
specified camera trajectories. Separately implemented checks assess
geometric and mechanical consistency; ``independent'' here describes the check
implementation, not validation by an external laboratory. Reconstruction and optical-flow
studies then evaluate the resource under their respective input contracts;
the reconstruction studies use the two COLMAP builds described in
Section~\ref{sec:pipeline-audit}.

\subsection{Population, geometry and acquisition}
The downloadable dataset stores the baseline bank in \texttt{benchmark/}.
Its subdirectories \texttt{v1}--\texttt{v18} are condition-package identifiers,
not successive releases. They enumerate the six families f01--f06 crossed with static,
FE-derived and prescribed non-FE motion. For example, f01 uses v1, v2 and v3;
v2 is its FE condition, not a second dataset.

\begin{table}[tbp]
\centering\manuscripttablestyle
\begin{tabularx}{\linewidth}{@{}lYclll@{}}
\toprule Family & Centerline & Motion scale & Static / FE / non-FE & Original role & Appearance \\
\midrule
f01 & Straight & $1\times$ & v1 / v2 / v3 & Development & Yes \\
f02 & Right turn, $30^\circ$ & $1\times$ & v4 / v5 / v6 & Development & No \\
f03 & Upward turn, $35^\circ$ & $2\times$ & v7 / v8 / v9 & Development & No \\
f04 & Left $27^\circ$, down $20^\circ$ & $1\times$ & v10 / v11 / v12 & Validation & Yes \\
f05 & Right turn, $38^\circ$ & $1\times$ & v13 / v14 / v15 & Recovery test & Yes \\
f06 & Up $25^\circ$, left $25^\circ$ & $2\times$ & v16 / v17 / v18 & Recovery test & Yes \\
\bottomrule
\end{tabularx}
\caption{Complete family/package map. Turns describe the configured centerline;
seeded wall perturbations, texture, camera offset and angular load locations
also vary by family. The motion multiplier scales all three spatial bases
within each motion class (Section~\ref{sec:mechanics}); it is not the peak
composed displacement. Every row supplies seven baseline streams: one rigid
reference and three motion conditions observed by two camera modes, or
630 images per family. Each ``Yes'' adds five 90-frame appearance streams of
the moving FE action, or 450 images, without adding a new anatomy. Roles are
original design assignments: all six families have been inspected for
dataset generation/reconstruction quality checks and are not unseen reconstruction tests.}
\label{tab:population}
\end{table}

Each surface has 48 circumferential by 112 longitudinal samples (5,376
vertices). We use $R$ for the median reference radius. Seeded radius perturbations add local relief to the straight or
piecewise-curved centerline (Figure~\ref{fig:atlas}). Terminal rings are
closed by fan triangles, an artificial closure rather than anatomical
material. The six engineered families do not sample a clinical population.

A family first supplies an undeformed \emph{reference scan} with a moving
camera. Its three \emph{action conditions} then hold the surface static,
apply FE-derived motion, or apply a directly prescribed non-FE field.
Each action is observed with a moving camera and with a stationary camera,
using identical material states. Thus camera mode and surface-motion condition
are crossed in the baseline within every family. The appearance extension is
a separate restricted comparison: four families, moving cameras and FE actions
only; it does not cross all camera, motion and appearance factors.
Six shared references plus
$6\times3\times2=36$ actions give 42 unique streams, each with 90 images at
$640\times480$ pixels and 15\,Hz: 3,780 images. Shared references are linked,
not counted again inside each package.

The stationary action uses the last reference pose. The moving action follows
a separate trajectory after repositioning; reference and action are distinct
acquisitions, and local reconstruction windows must not cross their join.
A stationary camera removes camera-induced flow but does not remove surface
motion. With no camera translation to provide conventional triangulation
baseline, stationary streams are not used for separate local SfM reconstructions. The reference scan and estimated moving-action global map therefore
serve different purposes; neither should be confused with the FE reference
mesh defined next.

\begin{figure}[tbp]
\centering\includegraphics[width=\linewidth]{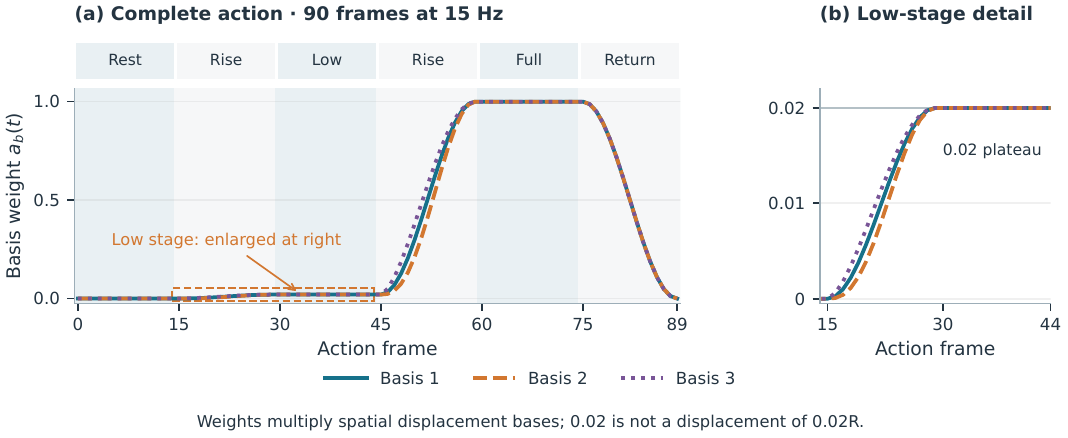}
\caption{Stored f01 weights for the three displacement bases. (a) The complete
action alternates rest, two rises and two plateaux before returning to rest.
The phase strip identifies the six 15-frame intervals, and the orange box
locates the enlarged region. (b) The first rise and low plateau expose the
weight of 0.02 and the three different transition profiles. A weight multiplies a vector-valued spatial basis; it is
not displacement divided by radius. FE and non-FE actions use this schedule,
whereas static actions apply zero displacement. Frames are action-local.}
\label{fig:schedule}
\end{figure}

Each deforming action combines three fixed spatial displacement patterns,
called \emph{bases}, with dimensionless time weights $a_1(t),a_2(t),a_3(t)$.
Figure~\ref{fig:schedule} shows when each pattern is activated: rest at frames
0--14, rise at 15--29, low plateau at 30--44, rise at 45--59, full plateau at
60--74, then return to rest by frame 89. The low weight is 0.02 of a basis
amplitude, not a displacement of $0.02R$. Slightly different transition
profiles change the mixture during ramps. Static actions apply zero motion
throughout. The optical-flow comparison (Section~\ref{sec:flow-study}) uses
f01--f03; appearance and reconstruction studies use the four families marked
in Table~\ref{tab:population}. These are exposed usage studies, rather than
unseen tests of a deformation-recovery method.
For future learning or parameter selection, partition by complete families:
one possible declared allocation is f01--f03 for development, f04 for validation
and f05--f06 for final evaluation. Keep every motion condition, camera mode,
reference and appearance variant of a family in the same partition; random
frame splitting would leak shared geometry and motion. All six released
families have been inspected in the present QA or usage studies. This allocation
is therefore a reuse protocol, not a claim of an untouched test in this article;
strictly unexposed confirmation requires separately frozen, uninspected families.

\subsection{Mechanical and prescribed motion}
\label{sec:mechanics}
\begin{figure}[tbp]
\centering
\begin{subfigure}{.96\linewidth}
\centering\includegraphics[width=\linewidth]{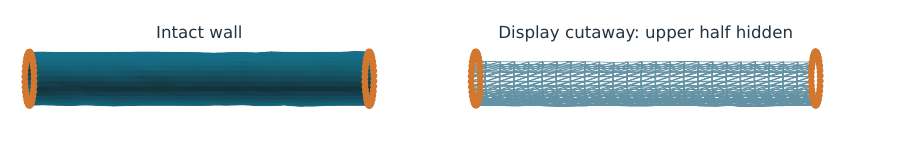}
\caption{FE wall: 1,600 nodes, 1,536 C3D6 elements. Orange marks the fixed end rings.}
\end{subfigure}\par\medskip
\begin{subfigure}[t]{.30\linewidth}
\centering\includegraphics[width=\linewidth,height=1.15in,keepaspectratio]{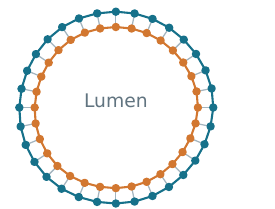}
\caption{Wall section: 32 inner and 32 outer nodes around the empty lumen.}
\end{subfigure}\hfill
\begin{subfigure}[t]{.30\linewidth}
\centering\includegraphics[width=\linewidth,height=1.15in,keepaspectratio]{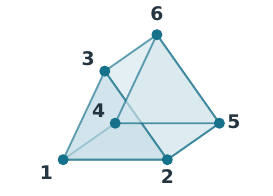}
\caption{Six-node prism topology: two triangular faces joined through the wall.}
\end{subfigure}\hfill
\begin{subfigure}[t]{.34\linewidth}
\centering\includegraphics[width=\linewidth,height=1.15in,keepaspectratio]{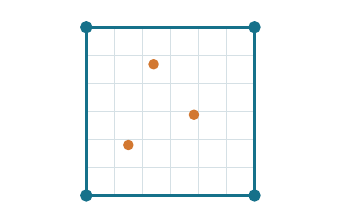}
\caption{Bilinear transfer from outer FE nodes (blue) to surface samples (orange).}
\end{subfigure}
\caption{From the mechanical wall to the rendering surface. The intact and
cutaway views show the stored f01 mesh; the cutaway hides its upper half for
viewing only. Element topology and interpolation are schematics. The
modeled wall has no longitudinal opening. Codex (version not recorded) assisted the topology and interpolation code.}
\label{fig:mechanical-geometry}
\end{figure}

Three representations must be distinguished. The \emph{reference surface}
is the undeformed textured boundary seen by the renderer. The \emph{FE
reference mesh} is a volumetric discretization of its hollow wall, used to
compute displacement bases. A \emph{reference scan} is an image acquisition
of the undeformed scene, not either mesh. Figure~\ref{fig:mechanical-geometry}
shows how the mechanical wall is discretized.

The FE wall has 24 axial intervals and 32 circumferential divisions, with
inner and outer nodes on 25 rings: $25\times32\times2=1,600$ nodes.
Each wall cell is split into two six-node triangular prisms (C3D6), giving
$24\times32\times2=1,536$ volume elements. Each prism has two triangular
faces joined by three edges through the wall; the lumen itself has no volume
elements. Nodes are mapped onto the family's geometry before assembling the
linear isotropic stiffness matrix $K$. Both terminal rings are fixed. All six families
use assigned Young's modulus $E=900{,}000$ and Poisson ratio $\nu=0.28$.
Lengths are in abstract scene units. These assigned parameters define a
controlled linear operator for reproducible load--response and energy queries.
They were not fitted to tissue measurements: constitutive calibration would
require measured stress--strain data, physical units and a validated tissue
model, outside this resource's controlled-kinematics objective.

For the FE condition, three localized smooth Gaussian load patches define
three separate load vectors $f_b$, $b=1,2,3$. Each produces a nodal
displacement basis $u_b$ by solving the same fixed-end system. With $f$ and
$c$ as subscripts denoting free and constrained degrees of freedom,
respectively,
\begin{equation}
 K_{ff}u_{b,f}=f_{b,f},\qquad u_{b,c}=0.
\end{equation}
Here $K_{ff}$ is the free/free stiffness block and $u_{b,c}=0$ enforces the
fixed ends. The response and load are scaled together to the assigned peak
surface displacement, preserving the linear equilibrium equation. The package stores node/element topology, fixed degrees of
freedom, the stiffness matrix in compressed sparse row (CSR) format, loads,
displacement, reactions and quadrature strain/stress. CSR stores matrix
entries row by row as three arrays: their values, column indices and pointers
to each row's start, avoiding a full dense matrix~\citep{scipycsr}. The stored
\texttt{K\_data}, \texttt{K\_indices}, \texttt{K\_indptr} and
\texttt{K\_shape} reconstruct $K$ directly. Outer-node responses are interpolated bilinearly
with periodic circumferential indexing onto the rendering surface. Therefore
the rendered material geometry and the generating FE volume are distinct
representations.

For time coefficients $a_b(t)$, the FE displacement is
$u(t)=\sum_b a_b(t)u_b$. Linear loads, reactions, strain and stress compose in
the same way. Total strain energy must retain cross terms:
\begin{equation}
 U(t)=\tfrac12 u(t)^T K u(t)
     =\tfrac12\sum_{b,c}a_b(t)a_c(t)u_b^T K u_c.
 \label{eq:energy}
\end{equation}
Here $U(t)$ is total strain energy; $u_b^T K u_c$ measures the interaction of
bases $b$ and $c$. Their deformed regions overlap, so the off-diagonal terms
need not vanish. Adding the three individual basis energies would generally
be incorrect. The evaluator
reader provides time-composed fields and total energy without an additional
per-frame mechanical archive. Voigt notation writes a symmetric
three-dimensional strain or stress tensor as six components~\citep{bleyervoigt}.
The exported strain order is
$(\epsilon_{xx},\epsilon_{yy},\epsilon_{zz},2\epsilon_{xy},
2\epsilon_{yz},2\epsilon_{zx})$, while stress uses
$(\sigma_{xx},\sigma_{yy},\sigma_{zz},\sigma_{xy},
\sigma_{yz},\sigma_{zx})$. Thus engineering shear strain is twice the
corresponding tensor entry; shear stress has no factor of two.

The prescribed non-FE condition moves the same rendering surface using three
Gaussian vector fields, combined with the same time weights and smoothly
tapered to zero at the ends. It provides exact kinematic truth without a
stiffness solve, constitutive model or equilibrium constraint.
The static condition instead leaves every surface vertex at its rest position.
For moderate families the three FE basis peaks are $(0.06,0.05,0.04)R$, whereas
the non-FE peaks are $(0.12,0.10,0.08)R$, with $R$ the median reference radius.
Both sets double for f03/f06. Thus amplitude is coupled to family and differs
between motion classes. Comparing conditions does not isolate the causal
effect of mechanical compatibility. The exact rendered displacement is
prescribed kinematic truth. Linear FE equilibrium certifies the declared
discrete model, while large displacement can violate its small-strain
assumption; this is a model limitation, not an unavoidable property of FE
analysis. Nonlinear FE can model other regimes but is not used here.

\subsection{Rendering and persistent material truth}
A pinhole camera with specified intrinsics uses
\begin{equation}
 p_c=R_{wc}^{T}(p_w-C_w),\qquad
 u=f_x p_{c,x}/p_{c,z}+c_x,\quad
 v=c_y-f_y p_{c,y}/p_{c,z}.
\end{equation}
Here $p_w$ is a world point, $C_w$ the camera center, and $R_{wc}$ maps camera
axes into world coordinates; $p_c$ is the resulting camera-frame point.
Focal lengths $(f_x,f_y)$ and principal point $(c_x,c_y)$ are in pixels.
The minus sign in $v$ converts the upward camera axis to downward image rows.
Camera axes are right/up/forward and pixel centers are
$(\mathrm{column}+0.5,\mathrm{row}+0.5)$. Depth is camera $z$, not Euclidean range.
Both synthetic renderers use the PINHOLE model, without lens distortion;
their calibration is provided per stream. The reproduction package specifies
numerical storage and projection precision.

\begin{figure}[tbp]
\centering
\includegraphics[width=\linewidth]{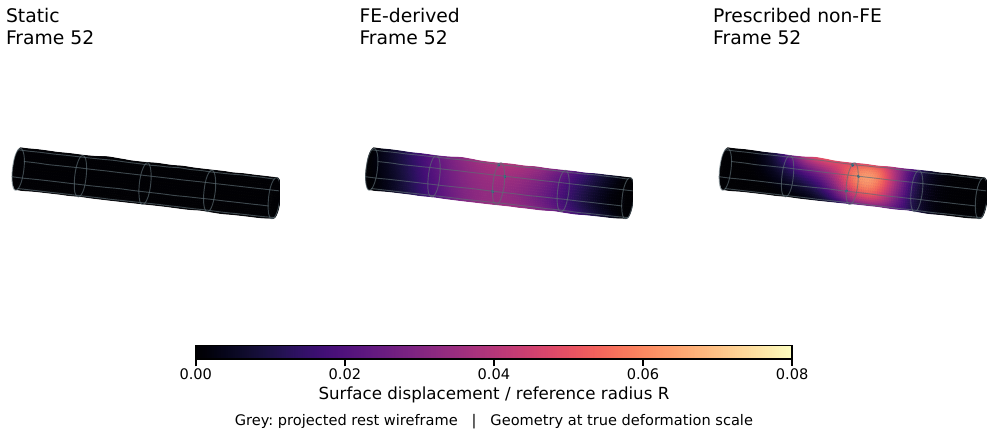}\par\medskip
\begin{subfigure}[t]{.32\linewidth}
\centering\includegraphics[width=\linewidth]{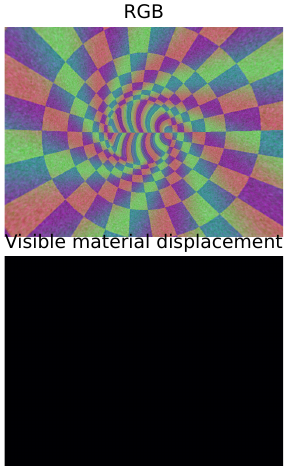}
\caption{Static wall}
\end{subfigure}\hfill
\begin{subfigure}[t]{.32\linewidth}
\centering\includegraphics[width=\linewidth]{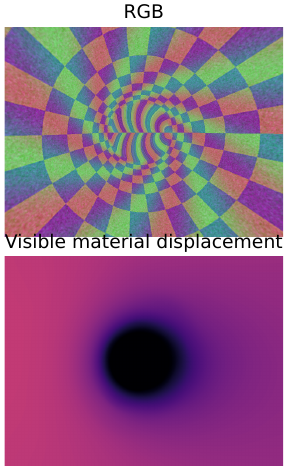}
\caption{FE-derived motion}
\end{subfigure}\hfill
\begin{subfigure}[t]{.32\linewidth}
\centering\includegraphics[width=\linewidth]{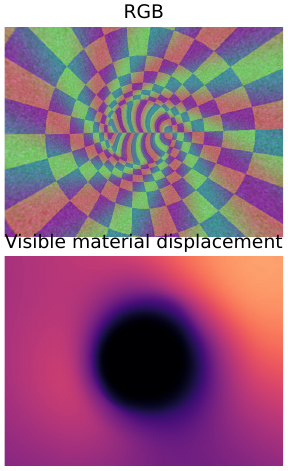}
\caption{Prescribed non-FE motion}
\end{subfigure}\par\smallskip
\includegraphics[width=.90\linewidth]{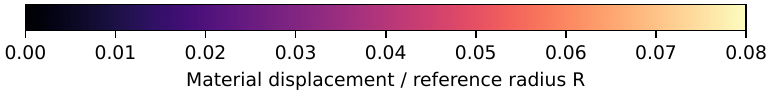}
\caption{Stored f01 motion at frame 52. The upper row shows the three-dimensional
surface at its true deformation scale, coloured by displacement magnitude divided
by reference radius $R$; colours average vertex magnitudes per face. Grey lines
project the rest wireframe. Below, the retained RGB and image-space material
maps use the same stationary camera; each foreground pixel measures its material
point's displacement from rest. Both colour scales span $0$--$0.08R$, without
motion magnification. Grey in the image maps denotes background; black denotes
zero displacement. FE and non-FE fields have different prescribed amplitudes.}
\label{fig:material}
\end{figure}

In the baseline renderer, RGB and dense truth share a perspective-correct
z-buffer. A valid pixel records
a frontmost triangle $q=(i_1,i_2,i_3)$ and physical barycentric weights
$\lambda=(\lambda_1,\lambda_2,\lambda_3)$, the fractional contribution of each
triangle vertex to the surface point ($\lambda_j\geq0$, $\sum_j\lambda_j=1$
up to numerical tolerance). At a vertex its weight is one; inside a triangle
the three weights interpolate the position. Its persistent material point is
\begin{equation}
 X_t(q,\lambda)=\sum_{j=1}^{3}\lambda_j V_t(i_j),\qquad
 \Delta X_{s,t}=X_t-X_s.
\end{equation}
Here $V_t(i_j)$ is vertex $i_j$ at frame $t$. The triangle identity $q$ and
weights $\lambda_j$ stay fixed while its vertices move; $\Delta X_{s,t}$ is
therefore the displacement of one material point between frames $s$ and $t$,
not the nearest-point difference between two surfaces. These are surface
material weights rather than affine image-space weights. They connect RGB pixels to reference and deformed geometry
(Figure~\ref{fig:material}), and support arbitrary sparse material queries.
World face normals follow triangle winding. Foreground has a true valid mask,
a nonnegative triangle index, positive finite camera depth and finite material
weights, positions and normals. Background has a false mask, triangle $-1$
and NaN geometric values, so it cannot be scored as a surface observation.

The baseline samples a fixed artificial checker texture at each material
location, without a physical lighting model. Angular interpolation unwraps
seam-crossing triangles before periodic nearest-texel sampling. Per-frame gain
jitter (uniform bounds of $\pm$0.5--1.8\% by family) and Gaussian noise (standard deviation
0.35--1.3 on the 8-bit channel scale) precede clipping and integer conversion.
Thus even a static scene under a fixed camera has photometric variation.
The appearance add-on below replaces this image formation with path tracing.
Before perspective projection, the current rasterizer clips triangles at the
camera near plane ($z=10^{-5}$). Each new polygon vertex carries material
weights on its original face, so fan triangulation retains the original face
identity and perspective-correct material interpolation. Preview and dense-truth
rendering use the same clipping rule. The sealed baseline uses its retained
producer: a complete census of all 3,780 views, checked by a separate implementation with
camera-frustum half spaces, finds no visible near-plane-crossing face. The
correction therefore preserves the released views while supporting crossing
triangles in new configurations.

Forward optical flow projects the same material point into the next camera
and subtracts its source pixel. A finite positive-depth projection defines
\emph{flow validity}, independently of target visibility. Continuous target-ray
surface queries distinguish visible, occluded, out-of-frame, behind-camera
and unresolved points. Visibility uses tolerance
$10^{-5}+10^{-5}|z|$ in scene units. The categories are mutually exclusive on
source-valid pixels. The final frame has no next-frame flow by definition;
backward and long-range flow are derived on demand from the same states and
material identities.

\subsection{Matched appearance and illumination extension}
The appearance add-on, stored as \texttt{appearance/} in the downloadable dataset, changes how an
existing scene looks while retaining its motion. It reuses only the moving
FE actions of the four families marked in Table~\ref{tab:population}: f01
(package v2), f04 (v11), f05 (v14) and f06 (v17). These are selections from
the baseline bank, not new versions or new anatomies. Each contributes the same 90
states under five conditions: diffuse symmetric lighting, diffuse asymmetric
lighting, coated symmetric lighting, and half/double exposure of the diffuse
symmetric radiance. Thus 360 shared truth states and 1,080 traced radiance maps
produce 1,800 RGB images. The compact design selects one originally designated
development family (f01), the validation family (f04), and the two recovery-test
families (f05/f06); it spans straight, singly bent and doubly bent geometry
and both amplitude settings. f02/f03 do not have appearance variants. This
bounded selection preserves the original roles and controls production cost;
it is not an exhaustive anatomy or curvature experiment. f01 remains development, f04 validation and f05/f06
retain their original final-test designation. All conditions were declared
before reconstruction; inspecting this experiment does not leave these scenes
unseen for subsequent reconstruction development.

Coordinates are uniformly converted to an assigned median radius of 15\,mm.
This chosen scale sets the relationship between tube size and the 3\,mm light
offsets and permits readable metric units. It is a synthetic scale assignment,
not a measurement of a patient's lumen or a calibration of tissue mechanics.
The actual float32 render vertices are stored in metres; source rotations are
projected onto proper rotations by singular-value decomposition and retained
in float64. This defines the appearance renderer's actual camera/geometry
truth. It does not convert the original abstract-unit FE parameters into
validated SI tissue mechanics. Topology and surface material identities are
preserved. Baseline raster labels are not reused as path-tracer labels.

The open-source physically based renderer Mitsuba 3.9.1~\citep{mitsuba3renderer}
renders the appearance extension. It uses Dr.Jit 1.5.0, a just-in-time
compiler for parallel rendering computations~\citep{jakob2022drjit}.
A two-sided surface is rendered at $640\times480$
with pinhole intrinsics $(f_x,f_y,c_x,c_y)=(455,452,320,240)$. Two point lights
lie 3\,mm to either side of the camera. Symmetric light intensities are
$(0.0005,0.0005)$; asymmetric intensities are $(0.00075,0.00025)$ in the
renderer configuration, preserving their sum. These are controlled numerical
intensities, not measured endoscope illumination. The original artificial
checker texture is decoded to linear RGB and sampled with periodic bilinear
UV interpolation. The diffuse control uses diffuse reflection; the coated
condition uses the renderer's rough-plastic GGX model with roughness 0.18 and
interior/exterior indices 1.38/1~\citep{mitsuba3bsdf}. This contrast changes
reflectance and brightness jointly, rather than isolating specularity alone.

\begin{figure}[tbp]
\centering\includegraphics[width=\linewidth]{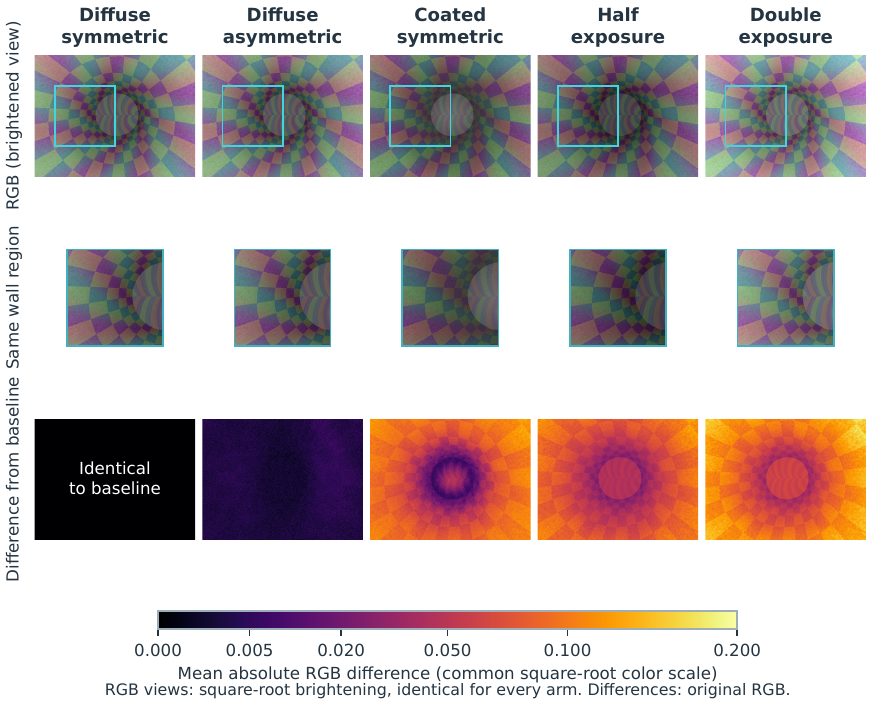}
\caption{Matched appearance at f01/frame 67. Each column shows one retained RGB
image and enlarges the same cyan-marked $240\times240$-pixel wall region
(rows 120--359, columns 80--319, zero-based). To reveal the darker wall, both
RGB rows use the identical display mapping $I_{\mathrm{display}}=\sqrt{I/255}$
on every saved encoded channel $I\in[0,255]$. This is a viewing adjustment,
not regenerated illumination; there is no per-image normalization. The bottom row makes changes relative to diffuse
symmetric visible: each pixel is the mean absolute difference of its three
original encoded RGB channels after division by 255, before display brightening.
A common square-root color scale
resolves subtle lighting changes without saturating any displayed value;
these are image differences, not physical radiance or reconstruction errors.
Geometry, camera and texture coordinates are identical across columns.
Lighting redistributes left/right power, coating changes reflectance, and
exposure applies a 0.5 or 2 gain to the baseline linear radiance before encoding.}
\label{fig:appearance}
\end{figure}

A bounded development screen fixed 256 samples per pixel, path depth four,
eight rendering threads and seed $(20260911+i)$, where $i$ is the zero-based
action-frame index (0--89). Pixel-center rays
have no spatial antialiasing or motion blur. Shared seeds match random-number
initialization, not identical transport samples across materials. Half/double
exposure applies gains 0.5/2 to shared linear radiance, followed by clipping,
sRGB encoding and 8-bit rounding; no adaptive exposure or sensor-noise model
is applied. Figure~\ref{fig:appearance} shows the fixed development frame 67.

Fresh dense records contain triangle ID, valid mask, camera-z depth and tracer
material barycentrics. Positions and geometric face normals are derived from
the shared states; float64 label storage does not imply float64 intersection
arithmetic. Cross-time material projection provides forward/in-frame support,
not occlusion-aware optical flow. Baseline flow/visibility capabilities must
therefore not be attributed automatically to this extension. Appearance
contrasts hold the inherited moving-camera FE action fixed; they do not
establish appearance interactions for stationary cameras, static actions or
non-FE motion.

\subsection{Data organization and inference/evaluation interfaces}
\label{sec:interfaces}
\label{sec:data-records}
Table~\ref{tab:records} separates sensor inputs, evaluator truth and estimated
reconstructions. The baseline has 3,780 dense maps and 3,738 forward-flow maps:
each 90-frame stream has 89 next-frame pairs. The package index resolves every
family, condition, acquisition and local reconstruction window; relative
links share references without counting them as new observations. The
appearance add-on stores five RGB directories per family and shares its
geometry, tracer labels and radiance records. Its family directories are
\texttt{appearance/families/f01}, \texttt{f04}, \texttt{f05} and \texttt{f06};
condition-specific images are under each family's \texttt{public/} directory.

Checksum manifests identify dataset members and relative links. They distinguish
canonical observations, shared references, estimated reconstructions and
reproduction records; archive membership and storage totals are supplied with
the release, rather than treated as additional samples.
Original generation metadata retain their own identifiers and hashes. A
verified package mapping and a read-only reproduction view reconcile those
records with the public package layout without changing sensor or evaluator
payloads.

\noindent\begin{minipage}{\linewidth}
These interfaces are Python reader classes in LumenMorph's
\texttt{backend.src.benchmark\_reader}, not directory names. Each is
constructed with a condition-package path such as \texttt{benchmark/v2};
the software separates method inputs from evaluation truth:
\begin{itemize}
\item \texttt{PublicDataset} reads RGB images, calibration and timestamps
from that package's \texttt{public/} paths.
It never opens evaluator files, so an image-only algorithm cannot obtain
ground-truth geometry through this interface.
\item \texttt{EvaluationDataset} additionally opens the explicitly privileged
\texttt{evaluation/} paths and provides material-point trajectories,
displacement, flow and time-composed FE fields. Static actions and rigid
references return zero applied mechanical fields; non-FE actions reject
FE-mechanics queries because their motion is prescribed without equilibrium.
\end{itemize}
The appearance extension provides two corresponding readers,
\texttt{AppearancePublic} and \texttt{AppearanceEvaluation}, in the module
\path{backend.src.appearance_reader}. Each takes a family directory such as
\path{appearance/families/f01}.

\end{minipage}\par\medskip

\begin{table}[tbp]
\centering\manuscripttablestyle
\begin{tabularx}{\linewidth}{@{}>{\raggedright\arraybackslash}p{.28\linewidth}YY@{}}
\toprule Record & Content & Information boundary \\
\midrule
\texttt{public/} & RGB, intrinsics, timestamps, acquisition metadata, shared reference links & Sensor/acquisition inputs \\
\path{evaluation/<stream>/states.npz} & Vertices, displacements, camera transforms, time & Evaluator-only prescribed geometry \\
\path{evaluation/<stream>/dense/*.npz} & Depth, face ID, material barycentrics, normals, valid mask & Evaluator-only dense truth \\
\path{evaluation/<stream>/flow_forward/*.npz} & Pixel flow and visibility categories & Evaluator-only correspondence truth \\
\texttt{evaluation/family/} & FE nodes, elements, parameters, stiffness and response bases & Generating-model state, not inferred tissue \\
\texttt{reconstruction/}; \texttt{bundles\_moving/} & Estimated reference/action globals and local models; image/window mapping & Independently estimated additional inputs \\
Provenance/qualification & Source/environment records, checksums, generation and camera QA & Reproduction and interpretation \\
\bottomrule
\end{tabularx}
\caption{Records in a baseline condition package \texttt{benchmark/vN}
and their inference/evaluation roles. Paths are relative to that package;
\texttt{<stream>} denotes moving or stationary, and relative links provide
the shared family/reference assets. The collection contains 24 global and
162 local reconstructions in total, not in each package. Public images, calibration and clocks are separated from privileged geometry,
correspondence and mechanics. Dense and forward-flow records describe the
baseline bank; the appearance add-on provides fresh tracer labels and
cross-time projection, but not baseline-equivalent occlusion-aware flow.
Reconstructed models are estimated outputs, not substitutes for ground truth.
Lossy MP4 previews are illustrative; PNGs are the canonical images.}
\label{tab:records}
\end{table}

\begin{table}[tbp]
\centering\manuscripttablestyle
\begin{tabularx}{\linewidth}{@{}>{\raggedright\arraybackslash}p{.34\linewidth}YY@{}}
\toprule Annotation & Baseline bank & Appearance extension \\
\midrule
RGB, calibration, timestamps & Stored & Stored for five matched conditions \\
Camera poses, deformed vertices, topology & Stored & Stored; shared by appearance conditions \\
Depth, validity, triangle ID, material weights & Stored dense maps & Stored fresh tracer labels \\
World positions, face normals & Positions derived; normals stored & Derived by appearance reader \\
Material trajectories and displacement & Derived from identity and states & Derived from identity and metre-valued states \\
Next-frame flow and visibility categories & Stored; arbitrary stored-time pairs derivable & Material projection derivable; no equivalent occlusion classification \\
FE state, loads, stiffness, stress/strain & Stored bases; time fields composed for FE actions & Inherited family generating records; not fresh image-space mechanical labels \\
\bottomrule
\end{tabularx}
\caption{Annotation availability, separate from access permissions. ``Derived''
means computed from released records, not estimated from RGB. Non-FE actions
have prescribed kinematics without applied FE equilibrium; static/reference
fields are zero. Appearance coordinates use the declared render-world metre
scale and do not calibrate FE forces or tissue properties. The baseline CLI
scorer is not an automatic adapter for the appearance schema.}
\label{tab:availability}
\end{table}

``Evaluator-only'' denotes an interface and inference contract, not restricted
distribution: the privileged annotations are public and downloadable. Users can
open them with the evaluation readers, but must exclude them from image-only
inference. Table~\ref{tab:availability} distinguishes stored annotations from
derived quantities and prevents attributing baseline visibility to appearance
projections.

The prediction evaluator is a separate scoring utility in LumenMorph,
\texttt{backend.src.benchmark\_evaluate}, not a prediction method or RAFT.
After any method has produced a saved prediction array, this utility accepts
depth, next-frame flow,
oriented world normals or material displacement at native image resolution.
Submitted arrays have shape $H\times W$ for depth, $H\times W\times2$
for flow, and $H\times W\times3$ for world normals or displacement.
They are stored in an NPZ file with a \texttt{prediction} array and an
optional Boolean availability mask. The utility computes depth absolute
error, angular normal error and Euclidean flow/displacement endpoint error,
together with prediction coverage. RAFT supplies flow predictions in the
usage study; it does not construct the scorer or its ground truth. The eligible truth pixels
are fixed before predictions are inspected. An invalid or missing prediction
reduces coverage; error over the entire eligible population is available only
at full coverage. Otherwise the reported error is explicitly conditional on
the available predictions. For example, predicting only an easy wall region
cannot receive a full-population score.

\noindent\begin{minipage}{\linewidth}
This evaluator applies no truth-based scale fitting or alignment. An algorithm
must return the declared units and coordinate frame. SfM's arbitrary gauge is
handled separately by the camera-center similarity fits defined below.
Twelve regression tests and four constant-output controls verify the software
contract; scientific accuracy comes from the usage studies rather than these
software checks.
\end{minipage}\par\medskip

\paragraph{Evaluation tasks and outputs.}
For source pixel $p$ on the surface, let $m_p=(q_p,\lambda_p)$ denote its
recorded material identity. The linked truth is
$\{C_t,R_{wc,t},X_t(m_p)\}$; $m_p$ is used by the evaluator, not supplied
to image-only inference. Table~\ref{tab:tasks} defines reusable tasks.
Unless a known-information variant is explicitly declared, inputs are the
RGB sequence, calibration and timestamps from the public interface. Any
estimated reference model used additionally must be identified as such.
Freeze the method and predictions before opening evaluator records.

\begin{table}[tbp]
\centering\manuscripttablestyle
\begin{tabularx}{\linewidth}{@{}>{\raggedright\arraybackslash}p{.21\linewidth}YY@{}}
\toprule Task & Required prediction & Evaluation contract \\
\midrule
Per-frame geometry & Camera-z depth or oriented world normals at source pixels & Absolute depth / angular normal error and coverage \\
Image correspondence & Next-frame pixel displacement & Flow endpoint error and coverage on fixed visible support; separate visibility strata \\
Material motion & $\widehat{\Delta X}_{s,t}(p)$ for the material point visible at source $s$ & 3D endpoint error against $X_t(m_p)-X_s(m_p)$, with source-valid denominator \\
Complete deformable reconstruction & Estimated cameras and a time-indexed surface with predicted cross-time material associations & Camera errors, bidirectional geometry/support and material-motion error, each with coverage and declared gauge \\
Supplied-geometry interpolation & Displacement field from sparse observations, supplied geometry/identities and declared mechanical records & Displacement error under that privileged-input contract; not image-only reconstruction \\
\bottomrule
\end{tabularx}
\caption{Tasks enabled by the linked annotations. The dense CLI implements the
first three task rows for the baseline schema; surface and camera scoring use
the separate study tools. Complete deformable reconstruction is a defined
evaluation target, not a claim that the present studies solve it or that one
CLI command scores an arbitrary dynamic representation.}
\label{tab:tasks}
\end{table}

A complete reconstruction must predict associations across time rather than
select a new nearest surface point in every frame. A suitable joint target is
$\{\widehat C_t,\widehat R_{wc,t},\widehat X_t(a)\}_{t}$, where $a$ indexes
the method's persistent surface entities. Its declared output adapter must
associate these predictions with source observations, report missing
associations and evaluate their trajectories against $m_p$. A single gauge
must apply to the sequence and all predicted times; independently fitting each
state would obscure motion errors. Without external scale/orientation access,
report a declared gauge-invariant or explicitly truth-aligned score, rather
than absolute-world recovery. The existing dense scorer performs no such fit.
This protocol links camera, geometry and material-motion evaluation without
inventing a combined ranking or treating the current quasi-static SfM and
privileged interpolation studies as end-to-end deformable reconstruction.
A minimal reader/query/scoring example is given in Supplementary Methods S4.

\subsection{Extending the benchmark with LumenMorph}
\label{sec:extending-lumenmorph}
The fixed dataset and the generator are separately versioned resources.
LumenMorph \LumenMorphVersion{} exposes the staged producer
\path{backend.src.benchmark} and a JSON specification, exemplified by
\path{configs/tubular_benchmark_v1.json}. Within its supported tubular
parameterization, a user can configure seeds, centerline sections, radius,
surface sampling and bumps; camera intrinsics and path-progress/offset
parameters; load-patch locations, directions and widths; FE and prescribed
non-FE peak amplitudes; and the linear material and volume-mesh parameters.
Texture and photometric-noise settings control image generation. The separate
appearance producer supplies the matched tracing conditions described above.
The benchmark producer in this release retains its predefined 90-state motion
schedule; arbitrary temporal waveforms are not a configuration feature.

From a pinned checkout with the mechanics submodule initialized, users copy
and edit the specification, then invoke the producer's \texttt{freeze},
\texttt{prepare}, \texttt{package} and \texttt{render} stages in a fresh output
root. For example, \texttt{freeze --spec custom.json --root ./new\_suite}
records the chosen specification and sources before family preparation,
condition packaging and reference/moving/stationary rendering. The resulting
suite exports the linked RGB, geometry, material and mechanical records used
by the readers in Section~\ref{sec:interfaces}; generation commands and a
complete single-family example are given in Supplementary Methods S3.
Supplementary Methods S4 shows how to consume the fixed release without
generating another dataset.
Each new configuration requires its own consistency checks; validation of the
released benchmark does not automatically validate a newly generated scene.
The generic plane/tube preview CLI is a separate path and does not imply the
full mechanical benchmark schema.

\subsection{Reconstruction pipelines and reproducible configuration}
\label{sec:pipeline-audit}
Two source-traceable CPU builds test whether the resource can reveal
differences between a task-specific reconstruction pipeline and a maintained
general-purpose implementation:
\begin{itemize}
\item \textbf{Pipeline A: COLMAP-for-colonoscopy 2}, a derivative of the
colonoscopy adaptation introduced in Sub\'ias Sarrato's bachelor's
thesis~\citep{subias2021thesis,subiascolmap}, with subsequent developments
described in a master's thesis~\citep{cirauqui2022thesis}. Supplementary
Methods S1 preserves its inherited geometry/matching changes and source lineage.
\item \textbf{Pipeline B: a derivative of official COLMAP \OfficialCOLMAPVersion},
based on the upstream incremental SfM implementation~\citep{schonberger2016colmap}.
It provides a separately maintained reference with newer estimation,
matching and mapper implementations; Supplementary Methods S1 identifies the
execution-order corrections.
\end{itemize}
Both pipelines share COLMAP ancestry and use fixed public source releases
(Section~\ref{sec:code-availability}). They receive the same images,
calibration and requested windows, with disabled intrinsic refinement and
frozen thresholds. Their internal algorithms differ, so this is a benchmark
usage comparison, not an isolated algorithmic ablation. The declared profiles,
source repairs, fresh versus retained stages and runtime receipts are documented
in Supplementary Methods S1--S2. Retained stages are not fresh end-to-end
reproduction, and the supplied dataset baseline models have a separate
source/runtime identity from this corrected comparison.

\subsection{Cross-pipeline reconstruction experiment}
\label{sec:cross-pipeline-method}
SfM estimates camera poses and a sparse set of 3D feature points from multiple
images. A frame is \emph{registered} when the mapper estimates its pose;
registration alone says nothing about its accuracy. The frozen comparison
requests 20 family/condition combinations: the moving FE actions of four
families under five appearance conditions, with three computational repetitions
per pipeline (120 cases and 240 global/local reconstruction routes).
Fresh and retained inference stages follow the fixed contract in
Supplementary Methods S1--S2.
This reuses matching rather than claiming fresh end-to-end B reproduction.
Here \emph{global} means one reconstruction from all 90 consecutive
moving-action images, frames 0--89. \emph{Local} means a separately initialized
reconstruction from the 15 consecutive images at frames 60--74, the full-motion
plateau in Figure~\ref{fig:schedule}. Frame indices are zero-based and local
to the action acquisition; this is a fixed temporal window, not 15 randomly
selected frames or a subset cropped from the finished global model.
The local window deliberately holds the wall state constant while the camera
moves. It tests quasi-static reconstruction at a deformed state, rather than
tracking a changing surface; the global window spans the complete motion cycle.
Inputs are limited to images and calibration;
no ground-truth poses, geometry or material identities were supplied.
Each mapper receives a private copy of the post-matching database. Thread
counts, seeds and source derivatives follow Section~\ref{sec:pipeline-audit}.
Extraction, matching, global mapping and local mapping had respective time
caps of 300, 3,600, 3,600 and 1,200 seconds. This is the frozen parameter set
from the earlier appearance study, translated between pipelines as described
in Section~\ref{sec:pipeline-audit}.

\begin{figure}[tbp]
\centering
\includegraphics[width=\linewidth,height=.78\textheight,keepaspectratio]{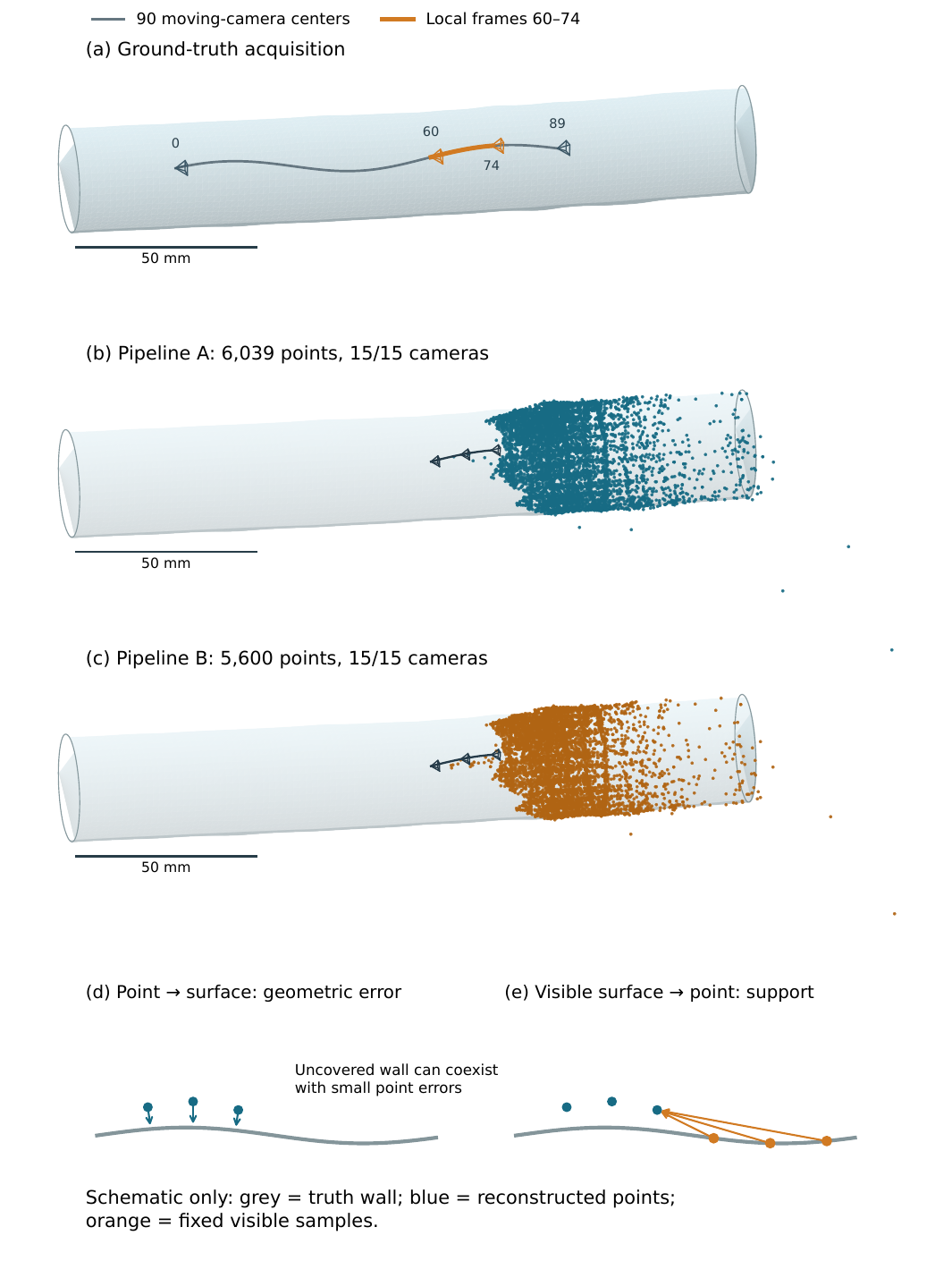}
\caption{Spatial interpretation of the reconstruction experiment.
(a) The ground-truth moving-camera trajectory and the local frames 60--74 inside
the f01 appearance wall at frame 67, which is constant over this plateau.
The near half of the wall is omitted for display; the complete mesh is used for
scoring. Camera glyphs indicate recorded positions/orientations, not calibrated
fields of view. (b,c) Retained A-SIFT1 and B-ORDER1 local reconstructions for the
diffuse-symmetric condition, repeat 1. Points and cameras use each model's recorded
center-fitted Sim(3), expressed in the same render-world basis; there is no
additional surface fit or absolute-scale recovery claim. Each pipeline has one
view retaining every point, including outliers, with identical A/B bounds and
scale. Scale bars use the assigned millimeter render-world scale.
(d,e) Schematic cross-sections distinguish point-to-surface error from the distance
of fixed visible-surface samples to reconstructed points; they contain no
experimental measurements. Codex (version not recorded) assisted the schematic code in (d,e).}
\label{fig:spatial-reconstruction}
\end{figure}

For evaluation, each estimated camera trajectory was aligned to ground truth
by a separately fitted proper similarity transformation (Sim(3): rotation,
translation and positive scale)~\citep{umeyama1991similarity}, using scored
camera centers only. This removes
the arbitrary coordinate system and scale of monocular SfM; the fitted errors
therefore do not measure absolute-scale recovery.
We write the fitted scale, rotation and translation as $(s,Q,t)$.
Synthetic truth is first expressed in the evaluator's COLMAP-compatible
world and camera bases: with $F=\operatorname{diag}(1,-1,1)$,
$C_{\mathrm{ref}}=FC_w$ and $R_{\mathrm{ref}}=FR_{wc}F$; geometry uses
the same world-basis change. This declared conversion is fixed before fitting.
C3VD's already-native poses receive no additional reflection.
Camera-position absolute trajectory error (ATE) is summarized by its root
mean square error (RMSE). For camera-to-world rotations $\hat R_i,R_i$,
absolute orientation error is the angle of $R_i^T Q\hat R_i$; relative
orientation compares $\hat R_j^T\hat R_i$ with $R_j^T R_i$ and needs no fitted
world gauge. Relative errors use adjacent requested frames only; a gap in
registration is not treated as a one-frame increment. Angles use an
\texttt{atan2} expression to retain resolution near zero.
For paired centers $\hat C_i,C_i$ and fitted similarity $(s,Q,t)$,
$\mathrm{ATE}_{\mathrm{RMSE}}=
\sqrt{N^{-1}\sum_i\|sQ\hat C_i+t-C_i\|^2}$.
At least three paired centers, nonstationary center spread and a
second/first centered singular-value ratio of $10^{-6}$ in both trajectories
are required; otherwise the alignment-dependent scores are unavailable.
Local
surface error is the unsigned distance from each retained reconstructed point
to the complete, constant truth mesh for frames 60--74. Surface support is
measured in the reverse direction, from a fixed visible-surface sample to
the nearest reconstructed point. The sample uses every 16th pixel in both
image axes across all 15 local frames, independently of registration success.
Thus point-to-surface error asks whether reconstructed points are close to
the wall, while surface-to-point support asks which visible wall regions have
nearby reconstructed points. Neither measures persistent material
correspondence by itself. A sparse, incomplete model can score well on the
first metric and poorly on the second: a few points on one accurately
reconstructed patch have low point-to-surface distance, while uncovered wall
regions remain far from every point. We report both distributions, point
counts and their tails; a median alone can hide large geometry outliers.
Neither nearest-surface direction establishes correct material identity.
Figure~\ref{fig:spatial-reconstruction} illustrates the acquisition window,
retained local camera/point models and the two metric directions. Its example
is fixed to f01/diffuse-symmetric/repeat 1 for illustration; quantitative
comparisons retain every requested family and appearance condition.

Original and corrected attempts are documented, including failures or missing
outputs. Complete final outputs are retained; the earlier raw-stage retention
boundary is stated in Section~\ref{sec:data-availability}.
Corrected source, options, cases and evaluators are fixed before replay;
no seed, initial pair or threshold is selected from accuracy outcomes.

\subsection{Repeatability experiment}
\label{sec:repeatability-method}
Three computational repetitions per family/condition and pipeline test
stability within a fixed source, runtime and CPU profile. Private database inputs,
full-precision outputs, stage completion and registered-frame identities are
checked; failures and absent models remain in the denominator. These are
repetitions of the same scenes, not independent observations or a portability
test. Supplementary Methods S1--S2 gives stage-reuse and hardware details.

Each returned model is scored using its own identifiable camera-center gauge.
B-minus-A and condition-minus-diffuse camera contrasts additionally use the
intersection of registered requested frames, fitting each model on that same
support. Unidentifiable gauges and unavailable model pairs remain explicit
abstentions. All 120 cross-derivative and 192 appearance contrasts are retained.
Families are reported separately; no significance test treats frames, points
or repetitions as independent acquisitions.

Scoring uses full-precision binary poses and verified exports. A separate
implementation fits the proper positive similarity by a quaternion
eigensystem~\citep{horn1987orientation}, checks rotations and computes
surface distances. Positive/negative controls and the full-cohort readback are
retained in the reproduction material. These agreements validate the scoring
calculation, not accuracy beyond the supplied reference.

\subsection{Provided-geometry mechanical-record consumer}
\label{sec:mechanical-consumer}
To demonstrate a concrete use of the FE records, we test displacement
interpolation with supplied rest geometry and material correspondence.
This task isolates the effect of an energy prior after observations have been
provided; it does not estimate correspondence or deformation from RGB.
We use the exposed families f01--f04, static frame 67 and FE/non-FE frames
37, 52 and 67: 28 fields spanning rest, low motion, a ramp and full motion.
Fifty-six surface vertices form the observed set: seven axial rings crossed
with eight circumferential positions. All methods receive the same noisy
Cartesian displacement observations and exact zero-displacement constraints
at the two terminal rings. Independent Gaussian noise has component standard
deviation $0$, $0.001R$ or $0.01R$, each under three fixed seeds. The resulting
252 cases are dependent synthetic conditions, not independent acquisitions.

The baselines are zero displacement and thin-plate-spline radial-basis
interpolation in normalized rest-world coordinates, using
\texttt{SciPy RBFInterpolator} with kernel $\phi(r)=r^2\log r$,
a degree-one polynomial and zero smoothing~\citep{scipyrbf}. Supplied fixed-end
vertices enter the interpolator with zero displacement. The FE consumer uses the
stored volume stiffness and periodic bilinear surface transfer, minimizing
\begin{equation}
 \|B_f x-y\|_2^2+\lambda x^T K_{ff}x/d,
 \qquad d=\operatorname{median}\operatorname{diag}(K_{ff}).
\end{equation}
Here $x$ contains the free nodal displacement components divided by $R$,
$y$ contains the normalized observations, and $B_f$ transfers free outer-node
displacements to the 56 observed material locations. Fixed ends are eliminated.
The two prespecified values $\lambda=10^{-4},10^{-2}$ probe weak and stronger
regularization after diagonal normalization. They are fixed examples, not
optimized settings, and both are reported;
neither is selected using heldout error. The generating response bases, loads,
coefficients, full displacements and motion labels are excluded from inference.
Providing the true stiffness and identities makes this a privileged-input
resource demonstration.

Manufactured tests verify the transfer, constrained solve and evaluator-truth
isolation. Observations and evaluator truth are sealed separately; predictions
are sealed before scoring. Endpoint error is evaluated on the 5,224 surface
vertices that are neither observed nor supplied fixed-end constraints. All
requested cases and methods are retained. The summaries average seeds and
time states within each family, then give families equal weight; static,
FE and non-FE motion and noise levels remain separate.

\subsection{Optical-flow methods and evaluation}
Optical flow estimates each visible pixel's two-dimensional displacement
between images. This tests the dense correspondence annotations under
separately controlled camera and surface motion. We compare a zero-motion
control, the classical Farneb\"ack and dense inverse search (DIS) methods
\citep{farneback2003polynomial,kroeger2016dis}, and RAFT~\citep{teed2020raft}, a
neural network trained on external data. ``Learned'' refers to RAFT's
pretrained estimator; no model is trained on this benchmark. We use the
\texttt{torchvision} implementation and its released pretrained
weights (\texttt{raft\_large}, \texttt{C\_T\_SKHT\_V2} recipe, SHA-256
verified)~\citep{torchvisionraft}, with no fine-tuning on any benchmark truth.
The fixed development cohort contains three families crossed with static,
FE and non-FE motion, moving and stationary cameras, and eight source frames
(0, 14, 22, 37, 52, 67, 82 and 88), each paired with its immediate successor:
144 image pairs. Selection uses the public suite index and verifies
family, condition and split rather than assuming package numbers.
Every pair is checked against its declared family, motion and acquisition
before inference and scoring.

All four methods use the same correctly identified pairs. Classical methods
receive grayscale and RAFT receives RGB; the released RAFT transform maps
uint8 input to $[-1,1]$, with 12 update iterations, original image dimensions
and no resizing. DIS uses its medium preset; Farneb\"ack parameters and the
exact checkpoint hash are fixed in the package. Three fresh processes run
the classical methods on one CPU thread and RAFT on an NVIDIA RTX 5070 Ti
Laptop GPU (12 GB), with seed zero and deterministic PyTorch algorithms.
RAFT uses IEEE float32, with TF32 and autocast disabled, deterministic cuBLAS
configuration and cuDNN benchmarking disabled; software and driver versions
are retained. Each batch loads a fresh RAFT model. The 1,728 requested
arrays are sealed before evaluator truth is read.

Endpoint error, its median and 95th percentile, and fractions above one and
three pixels are evaluated separately for visible, occluded and out-of-frame
source points. Finite truth defines the eligible denominator; missing or
nonfinite predictions reduce reported prediction coverage rather than
silently removing requested support. Source frames are averaged within each
family before equal-family aggregation, retaining condition, camera stream
and visibility stratum. Independent calculations check every pair's scores
and all aggregates, alongside byte equality of the three prediction batches.
These computational repetitions do not enlarge the development population.

\subsection{Static physical-phantom control}
\label{sec:c3vd-bridge}
Two public C3VD acquisitions provide static physical-phantom controls:
the 61-frame \texttt{trans\_t1\_a} sequence and two fixed 30-frame segments,
0--29 and 30--59, of \texttt{desc\_t4\_a}~\citep{bobrow2023c3vd}.
The segments share one acquisition and are not independent phantoms. A moving
clinical colonoscope observes a static physical wall; these controls test
image reconstruction and do not measure deformation recovery.

The control is prepared directly from all 121 original public color PNGs,
retaining the 0/0/30 source-frame offsets. Images are rectified into a
1350$\times$1080 PINHOLE camera with $f_x=f_y=700$, $c_x=675$ and
$c_y=540$ pixels. The original public calibration
is mapped into the same target PINHOLE camera using the authors' published
Python-loader ray convention~\citep{c3vdsource}. Native COLMAP target pixel centers are
$(\mathrm{column}+0.5,\mathrm{row}+0.5)$; source coordinates follow the
loader's declared principal-center and affine-stretch convention. The
alternative C++ matrix convention is distinguished in the provenance record.
The map keeps the optical axes fixed; no axis rotation is estimated from
reference poses or outcomes. A full-grid forward/inverse ray check verifies
the map, and a separate source/replay audit verifies every output PNG.
Interpolation and the published calibration remain physical approximations;
the numerical map tolerance is not a lens-accuracy claim.

The three fixed image sets and three repetitions of both corrected pipelines
request 18 global reconstruction routes. A uses A-SIFT1 and B uses B-ORDER1,
both with the CPU-AFFINITY1 profile. All nine A cases extract afresh and
execute fresh fixed-affinity mapping; matching follows the same exact-feature-only
inheritance rule as the appearance experiment. The nine B cases retain their
corresponding closed matching inputs and execute the fixed-affinity mapper.
Canonical payload equality, actual stage lineage and all three repetitions
are audited separately. Numeric thresholds, seeds, thread counts and stage limits
follow the synthetic comparison. No seed, initial pair, field of view or minimum model size is
selected to obtain full registration. All returned fragments and failures
remain in the requested denominator. Reference poses are reconciled against
the authors' column-major serialization and retained CSVs, using native
camera axes and the explicit source-frame offsets. Center and orientation
metrics follow the same identifiable proper-Sim(3) gauge rules; partial models
with insufficient geometric support receive an accuracy abstention. Position
errors are expressed in millimeters in the supplied C3VD coordinate system,
as supported by the source's translation and depth conventions~\citep{c3vdsource}, without
claiming metrological precision. Supplied C3VD poses are evaluator truth;
no known-pose model is used as image-based inference or as surface truth for
the rectified images.

\subsection{Visualization and AI assistance}
Data visualizations are generated by recorded Python scripts from retained
images, geometry, predictions and scores. Codex (OpenAI; client and model versions not recorded) assisted
analysis, validation and plotting code, including explanatory schematics. The executed scripts and their source/input receipts record
the reproducible procedures. Figure captions distinguish schematics from
measurements and disclose any display transformations. The author reviews
the code, results and manuscript and remains responsible for their content.

\section{Results}
\subsection{Geometry, image consistency and integrity}
Separately implemented generation checks distinguish continuous surface geometry from
the implementation's own raster output. A full geometric census covers all
1,161,216,000 pixels of the 3,780 saved frames. Separate sampled audits cover
119,616 flow vectors. Of 604,748 selected texture pixels, 519,638 are checked,
including 71,671 seam samples; 85,110 texel-boundary cases are skipped because
roundoff can change nearest-texel assignment there. They are not counted as
successful checks. Normals
are also sampled. Exhaustive geometry coverage must not be interpreted as an
exhaustive audit of every modality.

The retained criteria require relative physical-depth and material-point-depth
errors below $10^{-5}$, material reprojection error below $10^{-3}$ pixels,
and zero visibility mismatches. Shared-edge face ambiguity is accepted only
under the recorded barycentric/depth rule. The full census maxima are
$5.97\times10^{-8}$ for relative physical depth and $2.52\times10^{-4}$ pixels
for reprojection (rounded upward), with zero recorded depth or identity
failures. Sampled flow error is at most $3.71\times10^{-6}$ pixels against a
$10^{-3}$-pixel tolerance. Exact criteria and exception counts accompany the
machine-readable reports.

Periodic texture interpolation and higher-precision intermediate projection
support consistency across angular seams and grazing views. A complete readback verifies the retained files against their applicable
identity records and checks all relative links. Original generation metadata
and the public package layout reconcile exactly. These checks establish
artifact identity and synthetic consistency; they do not establish anatomical
realism or inverse-method accuracy.

Validation of the appearance extension by separately implemented checks accounts for all 360
shared states and verifies each consumed file against its declared identity. Pixelwise depth/projection
checks are complemented by 43,560 separately implemented sampled ray tests with zero
hit/miss disagreements and 43,560 cross-time material queries. All
radiance-to-RGB conversions reproduce exactly. Twelve fixed replays (frame 67,
four families, three traced conditions) reproduce RGB, radiance and geometry
exactly under the recorded environment. At near-tangent rays, the largest observed material-projection discrepancy is
$0.0047792$ pixels (41 pixels in 33 frames). The appearance release uses a
$0.01$-pixel precision tolerance, documented as a retrospective check of
released-label numerical precision rather than a prespecified test of
downstream reconstruction accuracy. The affected pixel/frame counts and full
projection audit accompany the release.

To isolate reconstruction jobs, the runner supplies each global and local
mapper a byte-identical copy
of frozen matches, checks unchanged canonical bytes, and protects
camera calibration, extracted image features and geometrically verified
matches in the COLMAP database; only auxiliary flag values and
physical SQLite layout may differ. The 120-job comparison fixes this
database-isolation rule in its execution protocol, with frozen parameters
and complete requested denominators; no model
is selected according to its accuracy.

\subsection{Discrete mechanics and source replay}
Independent arithmetic checks on 18 basis responses across all six FE families
give a maximum relative free-equilibrium residual of $1.77\times10^{-12}$ and
maximum relative integrated-energy error of $2.84\times10^{-13}$ (rounded
upward), against a $10^{-7}$ arithmetic tolerance. Strain/stress basis checks
also pass. Separate kinematic diagnostics cover all 90 composed states per
family and find positive quadrature-point deformation Jacobians.
These determinants measure local volume change in each sampled element:
positive values exclude a local inversion at those quadrature points. They
provide a discrete deformation sanity check, rather than a test of the
entire continuous surface.
Table~\ref{tab:mechanics} retains strain-regime diagnostics
rather than interpreting numerical equilibrium as physiological validity.
The relative Green--infinitesimal strain discrepancy reaches 0.145 in f03.
This is a kinematic diagnostic, not a measured tissue-model error. Positive
quadrature Jacobians do not prove global injectivity, self-contact avoidance
or continuum mesh convergence.

\begin{table}[tbp]
\centering\manuscripttablestyle\begin{tabular}{lrrr}\toprule
Family & $\max |\epsilon_{ij}|$ & $\max_t\rho_t$ & $\min\det F$ \\
\midrule
f01 & 0.00994 & 0.04022 & 0.99076 \\
f02 & 0.01087 & 0.03571 & 0.99193 \\
f03 & 0.02047 & 0.14501 & 0.98030 \\
f04 & 0.00876 & 0.04245 & 0.99109 \\
f05 & 0.01168 & 0.07836 & 0.98845 \\
f06 & 0.01912 & 0.08372 & 0.98323 \\
\bottomrule\end{tabular}

\caption{Minimum and maximum diagnostics over each retained 90-state FE audit. $\epsilon$ is the
infinitesimal strain tensor, $F=I+\nabla u$, and
$E_G=\tfrac12(F^TF-I)$. Here $\rho_t=\|E_G-\epsilon\|_F/
\max(\|\epsilon\|_F,10^{-30})$, with each norm taken over all elements,
Gauss points and tensor components at that time; the table reports
$\max_t\rho_t$, not a maximum pointwise ratio. The diagnostics characterize the declared small-strain model.}
\label{tab:mechanics}
\end{table}

The composed-mechanics reader was checked by a separate implementation on 27
static/FE/non-FE/reference calls. For a mixed-coefficient state in f01, total
energy is 374.86219 numerical units; the incorrectly diagonal-only sum is
354.0134. The full nine-term basis expansion agrees with
Equation~\ref{eq:energy}, demonstrating why derived access should preserve
mechanical cross terms instead of exporting potentially misleading scalar
sums.

A source-archive replay used relocated generator and solver source trees
outside Git. All 35 f01 preparation arrays match the sealed geometry, fields,
mechanics and poses exactly; the texture, first reference PNG and first-frame
dense maps also match. A second isolated environment freshly installed
pinned NumPy 2.4.4, SciPy 1.17.1 and OpenCV-Python 4.13.0.92 and reproduced
the same results, on the same Linux host; this is not a complete-bank
rerender or a cross-platform bitwise-identity guarantee.

\subsection{Utility of the mechanical records}
\label{sec:mechanical-consumer-results}
All 1,008 method/case outputs have finite predictions on the complete heldout
population. A separate evaluator reproduces 4,212 per-case and aggregate
arithmetic checks. The maximum FE normal-equation residual is
$1.15\times10^{-15}$. Table~\ref{tab:mechanical-consumer} shows that the FE
prior lowers equal-family mean heldout error for FE-compatible fields at all
three noise levels relative to geometric interpolation. At $0.001R$ noise,
the errors are $0.000984R$ and $0.000983R$ for the two FE priors, versus
$0.001224R$ for the radial-basis baseline.

On non-FE motion, both FE priors instead increase error at every noise level:
without noise, about $0.0083R$ versus $0.0016R$ for geometric interpolation.
With static noisy observations, all nonzero estimators produce false motion;
the zero-motion control has zero error. Thus the records enable a controlled
test of a mechanical assumption and its mismatch, rather than implying that
an FE prior must improve every field. The task supplies geometry, stiffness,
fixed ends and material identities; no practical image-based recovery or
clinical transfer conclusion follows from these results.

\begin{table}[tbp]
\centering\manuscripttablestyle\begin{tabular}{@{}llrrrr@{}}
\toprule Motion & Noise & Zero & RBF & FE $10^{-4}$ & FE $10^{-2}$ \\
\midrule
Static & 0.000 & 0.000000 & 0.000000 & 0.000000 & 0.000000 \\
Static & 0.001 & 0.000000 & 0.001090 & 0.000888 & 0.000886 \\
Static & 0.010 & 0.000000 & 0.010896 & 0.008879 & 0.008858 \\
FE & 0.000 & 0.032708 & 0.000278 & 0.000207 & 0.000208 \\
FE & 0.001 & 0.032708 & 0.001224 & 0.000984 & 0.000983 \\
FE & 0.010 & 0.032708 & 0.011442 & 0.009276 & 0.009255 \\
Non-FE & 0.000 & 0.020377 & 0.001617 & 0.008285 & 0.008307 \\
Non-FE & 0.001 & 0.020377 & 0.002290 & 0.008624 & 0.008643 \\
Non-FE & 0.010 & 0.020377 & 0.011717 & 0.014532 & 0.014530 \\
\bottomrule
\end{tabular}

\caption{Mean heldout material displacement endpoint error divided by $R$,
with equal family weighting over f01--f04. Noise is the observation component
standard deviation divided by $R$. Each FE/non-FE cell includes 36 requested
cases (four families, three times, three seeds); each static cell includes
12. Zero-noise repetitions share identical observations. All methods have
full finite coverage on 5,224 heldout vertices per case.}
\label{tab:mechanical-consumer}
\end{table}

\subsection{Baseline dataset: reconstruction coverage and camera error}
\label{sec:original-camera}
\begin{figure}[tbp]
\centering\includegraphics[width=\linewidth]{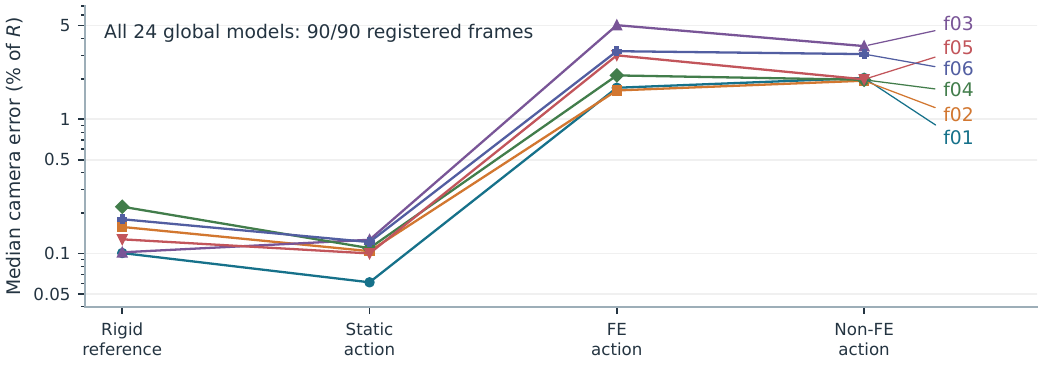}
\caption{Complete registration does not imply accurate cameras. Each marker is
one global reconstruction's median camera-center error after its own proper
similarity fit, expressed as a percentage of the family's reference radius
$R$ (logarithmic axis). Connected markers share a family across four
acquisitions; right-hand labels identify each line. All 24 models register
90/90 frames. The deformation conditions have larger errors, but differing
motion amplitudes and fields prevent a single-factor causal interpretation.
Table~\ref{tab:camera} gives the ranges in unscaled fractions of $R$.}
\label{fig:camera}
\end{figure}

Separately estimated reconstructions of the canonical 18-package bank use
the pinned dataset-generation CPU COLMAP-for-colonoscopy executable
(Section~\ref{sec:pipeline-audit}). All six rigid-reference globals and all
18 moving-action globals register their 90 images. Each moving action has nine
separately initialized 15-frame local reconstructions beginning at frames
0,10,20,30,40,50,60,70,75;
all 162 locals register every required frame. Stationary actions have zero
camera baseline and are excluded from these separate local reconstructions. For evaluation only, each camera-center trajectory is aligned
separately to truth with a proper rotation, translation and
scale~\citep{umeyama1991similarity}, and errors are normalized by the median
true reference radius. Table~\ref{tab:camera} and Figure~\ref{fig:camera}
retain the observed global camera bias despite full coverage; independent
local similarity fits remove each window's gauge and cannot validate
global/local alignment or absolute-scale recovery. These metrics describe
the distributed baseline reconstructions, whose source/runtime differs from
the corrected execution profiles. The controlled two-pipeline comparison is
reported separately in Section~\ref{sec:two-pipeline-results}.

\begin{table}[tbp]
\centering\manuscripttablestyle\begin{tabular}{lrr}\toprule
Acquisition & Min. sequence median / $R$ & Max. sequence median / $R$ \\
\midrule
Rigid reference & 0.001006 & 0.002230 \\
Static action & 0.000608 & 0.001263 \\
FE action & 0.016414 & 0.050363 \\
Non-FE action & 0.019402 & 0.035162 \\
\bottomrule\end{tabular}

\caption{Ranges of per-sequence global median camera-position errors across
six families, after separate proper Sim(3) evaluation fits. Errors are divided
by each family's median reference radius $R$; each column is the minimum or
maximum of six sequence medians, not a pooled point/camera percentile or an
uncertainty interval. Every global model registers 90/90 frames. The rigid
reference and static moving action are separate acquisitions.}
\label{tab:camera}
\end{table}

\subsection{Cross-pipeline reconstruction on synthetic and phantom images}
\label{sec:two-pipeline-results}
The matched comparison distinguishes registration coverage, camera-position
error, sparse point count, geometric error and visible-wall support. Each
quantity is evaluated on its declared support; common-frame contrasts are
reported separately from each model's own trajectory fit. Missing models,
fragments and unidentifiable gauges remain in the requested population.

\begin{figure}[tbp]
\centering
\begin{subfigure}[t]{0.49\linewidth}
\centering\includegraphics[width=\linewidth]{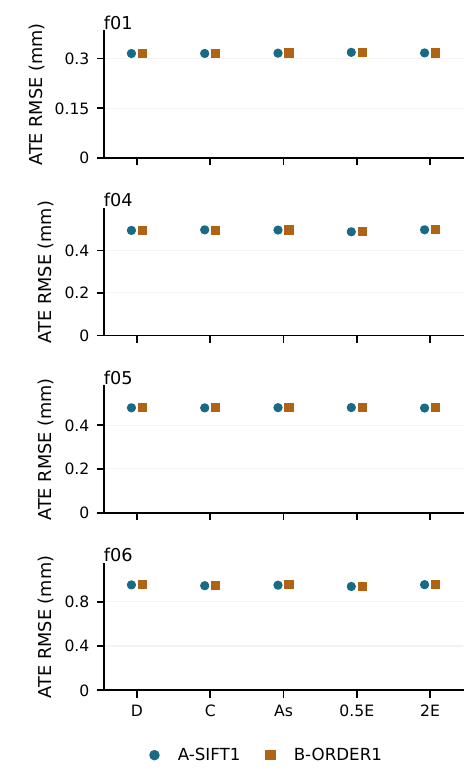}
\caption{Global: action frames 0--89.}
\end{subfigure}\hfill
\begin{subfigure}[t]{0.49\linewidth}
\centering\includegraphics[width=\linewidth]{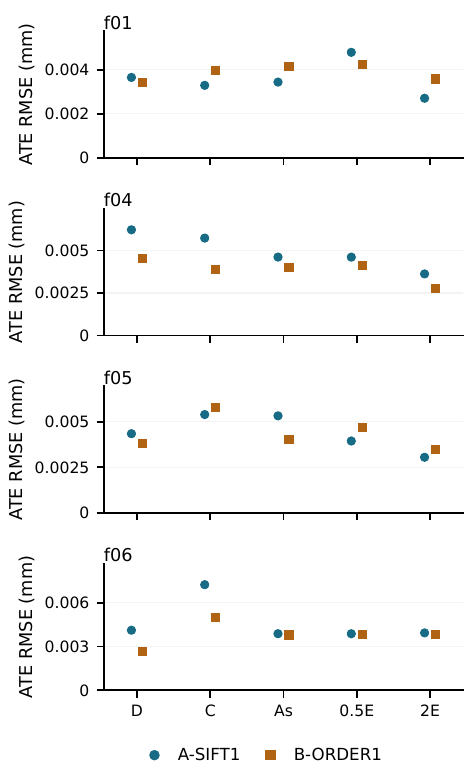}
\caption{Local: action frames 60--74.}
\end{subfigure}
\caption{Camera-center ATE RMSE under each model's own evaluation similarity
fit, in millimeters. D: diffuse symmetric; C: coated symmetric; As: diffuse
asymmetric; 0.5E and 2E: half and double exposure of the diffuse-symmetric
condition. Global and local denote independently reconstructed 90-frame
and 15-consecutive-frame trajectories, respectively
(Section~\ref{sec:cross-pipeline-method}); indices are action-local. Each family
has its own vertical limits with a zero origin. Markers summarize three computational repetitions; their min--max ranges
are zero. Every model registers its complete requested trajectory.}
\label{fig:current-camera}
\end{figure}

All 120 synthetic jobs return one 90/90-frame global model and one
15/15-frame local model, giving 240 evaluated routes. The diffuse-symmetric
baseline is reported in Table~\ref{tab:primary-camera-current};
Figure~\ref{fig:current-camera} includes all five appearance conditions.
Across the four families and five conditions, global ATE RMSE spans
0.3154--0.9535\,mm for A and 0.3145--0.9525\,mm for B, whereas local ATE
spans 0.00271--0.00726\,mm and 0.00264--0.00579\,mm, respectively.
The local trajectories receive their own evaluation similarity fit, so this
difference does not establish practical recovery of the global camera bias.
On common requested frames, B-minus-A ATE differences span
$-3.88$ to $2.68\,\mu$m globally and $-2.24$ to $0.875\,\mu$m locally
(Supplementary Figure S1). B has lower ATE in 8/20
global and 14/20 local comparisons. These are descriptive comparisons of
the declared source/runtime profiles, not an isolated mapper ablation.

\begin{table}[tbp]
\centering\manuscripttablestyle
\begin{tabular}{lllllll}
\toprule
Family & Model & A coverage & A ATE (mm) & B coverage & B ATE (mm) & $B-A$ (mm) \\
\midrule
f01 & global & 90/90 & 0.3154 & 90/90 & 0.3156 & 0.0001997 \\
f01 & local & 15/15 & 0.003651 & 15/15 & 0.003418 & -0.0002324 \\
f04 & global & 90/90 & 0.4944 & 90/90 & 0.4947 & 0.0002967 \\
f04 & local & 15/15 & 0.006227 & 15/15 & 0.004536 & -0.001691 \\
f05 & global & 90/90 & 0.4789 & 90/90 & 0.48 & 0.001186 \\
f05 & local & 15/15 & 0.00434 & 15/15 & 0.003798 & -0.0005421 \\
f06 & global & 90/90 & 0.9514 & 90/90 & 0.9516 & 0.0002333 \\
f06 & local & 15/15 & 0.004127 & 15/15 & 0.002638 & -0.001488 \\
\bottomrule
\end{tabular}

\caption{Diffuse-symmetric synthetic baseline. A and B denote A-SIFT1 and
the B-ORDER1 mapper with its declared inherited matching inputs. ATE is
camera-center RMSE in millimeters after each model's own proper similarity
fit. Global uses moving-action frames 0--89; local uses the consecutive
frames 60--74, independently reconstructed as defined in
Section~\ref{sec:cross-pipeline-method}. Global spans changing wall states;
local observes a constant deformed state. Their temporal supports and separate
similarity fits differ, so the global/local errors do not isolate the benefit
of local reconstruction. The final column fits both models on
the same requested frames.
Entries summarize three computational repetitions, whose values coincide
exactly; they are not independent scenes or uncertainty intervals.}
\label{tab:primary-camera-current}
\end{table}

The static physical-phantom controls use the separately checked
source-calibrated inputs described in Section~\ref{sec:c3vd-bridge}.
All 18 requested routes are evaluated, with one returned fragment per route
(Table~\ref{tab:physical-current}). Both pipelines register 30/30 frames in
each descending subset. In the 61-frame transverse acquisition, each registers
only 2/61 frames; the two-center gauge cannot identify the declared similarity
fit, so absolute center and orientation accuracy remain unavailable.
On the complete descending subsets, center RMSE is 0.0293--0.0830\,mm,
while mean absolute orientation error is 10.2--28.7 degrees. Small fitted
center errors therefore do not by themselves demonstrate accurate camera
orientations. Source calibration does not restore transverse registration
under these fixed profiles. These static acquisitions contain no deformation
truth and do not establish clinical recovery.

\begin{table}[tbp]
\centering\manuscripttablestyle
\begin{tabular}{llllll}
\toprule
Case & Pipeline & Models & Max. coverage & ATE (mm) & Abs. rot. ($^\circ$) \\
\midrule
trans 0--60 & A & 1 & 2/61 & \textemdash & \textemdash \\
trans 0--60 & B & 1 & 2/61 & \textemdash & \textemdash \\
desc 0--29 & A & 1 & 30/30 & 0.08297 & 28.72 \\
desc 0--29 & B & 1 & 30/30 & 0.08298 & 27.16 \\
desc 30--59 & A & 1 & 30/30 & 0.02939 & 10.55 \\
desc 30--59 & B & 1 & 30/30 & 0.02926 & 10.16 \\
\bottomrule
\end{tabular}

\caption{Complete source-calibrated static-phantom controls. The public
color images and published calibration are remapped as described in
Section~\ref{sec:c3vd-bridge}. Coverage is the largest registered requested-frame count in
any returned fragment; every returned fragment is counted in Models.
ATE is center RMSE in millimeters, and Abs. rot. is mean absolute angular
error in degrees, after the proper similarity fit to camera centers
without an extra camera-axis offset. A dash denotes an unidentifiable
gauge. Full registration of the descending subsets coexists with mean
orientation errors of 10.2--28.7 degrees; registration and small fitted center
errors do not establish accurate camera orientations.
Each row summarizes three exactly agreeing computational repetitions.}
\label{tab:physical-current}
\end{table}

\subsection{Repeatability of cameras, point counts and surface support}
\label{sec:repeatability-results}
The complete requested design comprises 40 family/condition/derivative groups,
with three computational repetitions per group. These repetitions test computational stability on the same scenes;
coverage and reconstruction accuracy are evaluated separately.
All 40 synthetic groups and all six source-calibrated physical groups
pass the complete exact repeatability checks under the declared source/runtime/CPU
profile. The separately implemented numerical audit verifies all 258 reported routes and
all 312 synthetic camera contrasts, including physical accuracy abstentions.
All returned model pose exports preserve their binary values exactly.

\begin{figure}[tbp]
\centering
\begin{subfigure}[t]{0.32\linewidth}
\centering\includegraphics[width=\linewidth]{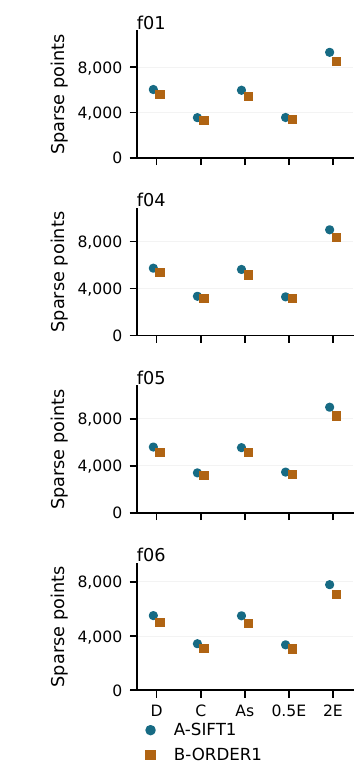}
\caption{Point count.}
\end{subfigure}\hfill
\begin{subfigure}[t]{0.32\linewidth}
\centering\includegraphics[width=\linewidth]{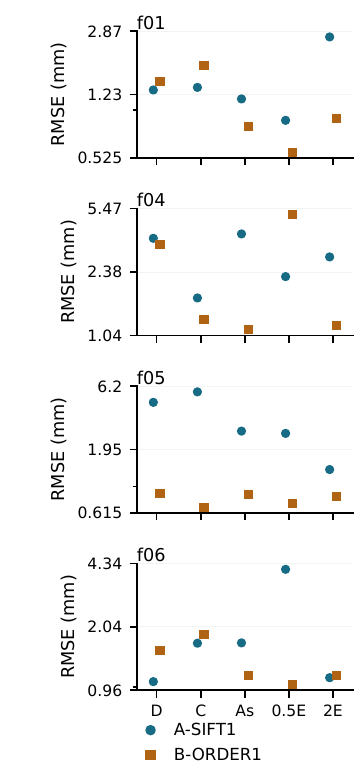}
\caption{Surface RMSE.}
\end{subfigure}\hfill
\begin{subfigure}[t]{0.32\linewidth}
\centering\includegraphics[width=\linewidth]{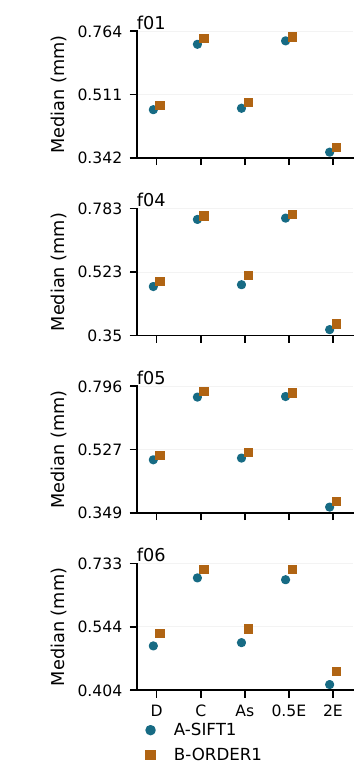}
\caption{Visible-wall support.}
\end{subfigure}
\caption{Local point counts, geometric error and support answer different
questions. Distance panels show millimeters on a logarithmic scale above
0.001\,mm, with a linear region below it; counts use a linear scale from
zero. Vertical limits vary by family. Condition codes follow
Figure~\ref{fig:current-camera}; three-repeat min--max ranges are zero.
Table~\ref{tab:appearance-local-current} gives every numerical value;
the accompanying metric records additionally retain medians and outlier tails.}
\label{fig:current-local-geometry}
\end{figure}

Repeatability does not ensure accuracy: the geometry/support results below
retain bias, missing wall regions and large outliers despite identical
fixed-profile outputs.
Table~\ref{tab:appearance-local-current} and
Figure~\ref{fig:current-local-geometry} retain every appearance condition.
A produces 3,285--9,319 local points and B produces 3,053--8,544. Point
counts are not normalized by surface area and do not imply uniformly dense
wall coverage. A has more points and a smaller median fixed-support
nearest-point distance in all 20 matched conditions; B has smaller
point-to-surface RMSE in 14/20.
Across both pipelines, median point-to-surface distance spans
0.0413--0.121\,mm, while RMSE spans 0.567--5.58\,mm. Retained outliers
reach 283\,mm for A in f05 coated-symmetric and 239\,mm for B in f04
diffuse-symmetric. The corresponding 95th-percentile distances are only
0.542 and 0.424\,mm, respectively. Typical proximity, sparse coverage and
outlier sensitivity thus require separate summaries.

\begin{table}[tbp]
\centering\manuscripttablestyle
\begin{tabular}{llllllll}
\toprule
Family & Condition & A points & B points & A RMSE & B RMSE & A support & B support \\
\midrule
f01 & D & 6039 & 5600 & 1.306 & 1.462 & 0.4641 & 0.4754 \\
f01 & C & 3554 & 3337 & 1.352 & 1.806 & 0.7033 & 0.7303 \\
f01 & As & 5973 & 5450 & 1.158 & 0.7982 & 0.4684 & 0.4853 \\
f01 & 0.5E & 3569 & 3378 & 0.8694 & 0.5672 & 0.7187 & 0.7367 \\
f01 & 2E & 9319 & 8544 & 2.657 & 0.8892 & 0.3543 & 0.3648 \\
f04 & D & 5726 & 5328 & 3.71 & 3.412 & 0.4774 & 0.4922 \\
f04 & C & 3335 & 3164 & 1.694 & 1.271 & 0.7309 & 0.7474 \\
f04 & As & 5618 & 5144 & 3.928 & 1.118 & 0.4828 & 0.5132 \\
f04 & 0.5E & 3285 & 3110 & 2.243 & 5.076 & 0.7371 & 0.7546 \\
f04 & 2E & 9000 & 8375 & 2.904 & 1.18 & 0.363 & 0.3764 \\
f05 & D & 5601 & 5159 & 4.614 & 0.8797 & 0.4927 & 0.5071 \\
f05 & C & 3402 & 3185 & 5.585 & 0.683 & 0.7411 & 0.7667 \\
f05 & As & 5548 & 5114 & 2.73 & 0.8544 & 0.4982 & 0.5176 \\
f05 & 0.5E & 3469 & 3262 & 2.618 & 0.7319 & 0.7438 & 0.7606 \\
f05 & 2E & 8988 & 8244 & 1.354 & 0.8236 & 0.3621 & 0.376 \\
f06 & D & 5513 & 5001 & 1.064 & 1.545 & 0.4978 & 0.5277 \\
f06 & C & 3436 & 3080 & 1.681 & 1.863 & 0.6848 & 0.713 \\
f06 & As & 5499 & 4935 & 1.688 & 1.151 & 0.5055 & 0.5391 \\
f06 & 0.5E & 3357 & 3053 & 4.051 & 1.028 & 0.679 & 0.7116 \\
f06 & 2E & 7795 & 7049 & 1.116 & 1.148 & 0.4153 & 0.4419 \\
\bottomrule
\end{tabular}

\caption{Complete local appearance comparison. Condition codes follow
Figure~\ref{fig:current-camera}. Point counts retain all reconstructed
points. RMSE is point-to-surface distance; Support is the median distance
from the fixed visible-wall support to its nearest reconstructed point.
Both distances are in millimeters and lower is better. Each value is the
median of three exactly agreeing computational repetitions, not an
uncertainty estimate. All 15 requested local frames register in every case.}
\label{tab:appearance-local-current}
\end{table}

\subsection{Optical-flow comparison}
\label{sec:flow-study}
Before evaluation, the protocol fixed source frames 0, 14, 22, 37, 52, 67,
82 and 88 and their immediate successors from both action acquisitions of
development packages v1--v9. This gives 144 pairs across three scene families. Zero flow, OpenCV Farneb\"ack
\citep{farneback2003polynomial}, OpenCV DIS-medium~\citep{kroeger2016dis} and
RAFT-large~\citep{teed2020raft,torchvisionraft} receive only the original
grayscale (classical methods) or RGB (RAFT) image pair; RAFT uses the fixed
GPU float32 profile and the released \texttt{torchvision} weights without benchmark-specific
fine-tuning.
Endpoint error is the Euclidean distance between predicted and true 2D flow
vectors, in pixels. It is measured on source-valid, finite-projecting, target-visible
pixels, with the same eligibility masks, denominators and per-pair/per-family
averaging for all four methods.

Three fresh full prediction batches use the same verified pair identities. All 1,728
requested arrays were finite, with no failed method calls and identical
prediction bytes and score rows across repeats. A separately implemented
scorer recomputed the errors, visibility-stratum support and aggregates,
agreeing on all 5,184 pair/method/stratum comparisons, including coverage and
empty-stratum checks. This verifies the scoring arithmetic independently of
the prediction methods.

Table~\ref{tab:flow-current} reports visible-pixel endpoint error with equal
family weighting. For moving cameras, RAFT has the lowest mean error in each
motion condition (0.188--0.189 pixels), followed by DIS (0.208--0.209 pixels).
For stationary cameras with deforming surfaces, DIS instead has the lowest
error: 0.0375 pixels for FE and 0.0408 pixels for non-FE motion, versus RAFT's
0.0474 and 0.0442 pixels. In the static stationary control, the zero predictor
has a numerical error of $1.13\times10^{-5}$ pixels. Its analytic flow is zero;
the tiny residual reflects numerical material reprojection, not camera or wall
motion. The image-based predictors
introduce small nonzero motion. Thus method ordering depends on the camera
and deformation condition, rather than supporting uniform RAFT superiority.

\begin{table}[tbp]
\centering\manuscripttablestyle\begin{tabular}{llllll}
\toprule
Motion & Camera & Zero & Farneb\"ack & DIS & RAFT \\
\midrule
static & moving & 9.378 & 0.3954 & 0.2082 & 0.1878 \\
static & stationary & 1.126e-05 & 0.01962 & 0.0208 & 0.008854 \\
fem & moving & 9.385 & 0.3936 & 0.2084 & 0.1883 \\
fem & stationary & 0.6793 & 0.05887 & 0.0375 & 0.04739 \\
nonfem & moving & 9.261 & 0.3928 & 0.2086 & 0.1893 \\
nonfem & stationary & 0.8128 & 0.0637 & 0.04083 & 0.04416 \\
\bottomrule
\end{tabular}

\caption{Visible-pixel mean endpoint error (pixels) on the corrected matched
144-pair development cohort. Eight source pairs are averaged within each
family, followed by equal weighting of the three families. Three computational
repetitions are summarized by their median; their predictions and score rows
are exactly equal. Every visible eligible vector has a finite prediction.}
\label{tab:flow-current}
\end{table}

There are no eligible occluded vectors in this cohort, so it provides no
occlusion-error result. For out-of-frame vectors, all eight pairs per family
have support with moving cameras; RAFT's equal-family mean errors are
0.463, 0.463 and 0.468 pixels for static, FE and non-FE motion. Stationary
streams lack complete eight-pair out-of-frame support, so their corresponding
aggregate is unavailable rather than silently averaging only supported pairs.
The package retains every pair, stratum, support count and unavailable score.

\section{Discussion}
The resource links geometry, persistent material identities, motion and
appearance so that evaluation questions can be posed under known acquisition
conditions. Users can distinguish camera-induced disagreement from prescribed
surface motion, examine whether sparse geometry supports the visible wall,
and test a mechanical assumption with supplied material correspondence.
The fixed release makes these questions accessible without generating a new
dataset; LumenMorph supports extensions within its published configuration
interface. The three usage studies illustrate complementary uses of the same
linked annotations.

In the SfM study, complete registration does not establish accurate camera
poses or adequate surface reconstruction. Point-to-surface proximity and
visible-wall support answer different questions: small typical distances can
coexist with sparse support or severe outliers. All requested synthetic frames
register and the fixed-profile computational repeats agree exactly, yet global
and local camera errors differ markedly. A reconstructs more local points and
achieves a smaller median visible-surface-to-point distance in every matched
condition, while B has lower surface RMSE in most conditions. Neither profile
dominates all reported measures. Large retained outliers also prevent small
surface medians from establishing uniformly accurate geometry.

The optical-flow study shows how controlled camera and surface motion expose
condition-dependent performance. A learned estimator is not uniformly best
on these inputs. The static stationary arm fixes geometry and camera pose
while retaining photometric perturbations, helping distinguish residual image
variation from geometric motion.

The mechanical-record study tests an energy prior after supplying geometry,
stiffness and material correspondence. Improved interpolation for compatible
fields and deterioration for prescribed non-FE fields reveal the usefulness
and possible mismatch of that assumption. The motion classes have different
amplitudes, so this is not a causal isolation of mechanical compatibility.
The experiment is a privileged-input resource demonstration rather than
image-based recovery.

The physical-phantom controls complement the synthetic comparison by
retaining reconstruction failures, partial coverage and unidentifiable gauges
on real images. Calibration and preprocessing provenance qualify their
interpretation. Their role is to expose transfer limits under the fixed
profiles. Complete descending registration coexists with 10.2--28.7-degree
mean orientation errors, while transverse registration reaches only 2/61 frames.
These controls do not establish uniformly accurate camera reconstruction,
deformation recovery or a general ranking of the pipelines.

The primary comparison uses complete binary poses and precisely verified
exports. Repetitions
measure computational stability on exposed scenes, not independent anatomical
confirmation. The local experiment observes a fixed deformed wall state;
its small fitted errors do not establish continuous deformation tracking.

Data consistency, repeatability and score replay answer separate questions.
Geometric/mechanical checks establish agreement with the stated generating
model, not measured tissue behavior. Repeated executions assess computational
stability within the pinned source/runtime/CPU profile. Scoring retained
predictions verifies metric reproduction without rerunning their inference.
Neither the synthetic checks nor the bounded fresh-environment consumer test
establish identical end-to-end outputs across arbitrary machines, operating
systems or builds.
For reuse, comparisons must verify input identities, keep family variants
together, state which annotations enter inference and report failures and
support alongside error. Algorithmic causes require controlled ablations.

\section{Limitations}
The dataset contains six engineered scene families, not 18 independent
anatomies. Geometry, texture, camera trajectories and material parameters are
specified by the generator rather than sampled from a clinical population.
The FE model uses linear elasticity and infinitesimal strain; the relative
Green--infinitesimal strain discrepancy reaches 0.145 in the strong-deformation
regime. Numerical equilibrium and positive quadrature-point Jacobians do not
validate tissue mechanics or large-deformation constitutive behavior.
Deformation amplitude also differs between FE and non-FE conditions, so their
comparison cannot isolate mechanical compatibility. Although the baseline
crosses camera mode and surface-motion condition, the appearance extension
covers only four families and moving-camera FE actions. It is not a complete
factorial design, and its findings do not establish interactions across all
camera, motion and appearance combinations.

The physical controls evaluate global reconstructions of two static phantom
acquisitions. Source-calibrated inputs use the checked PINHOLE remapping of
published omnidirectional calibration, with the interpolation and
lens-model approximations described in Section~\ref{sec:c3vd-bridge}.
Repeatability covers two corrected source derivatives, four synthetic families,
five appearance conditions and three source-calibrated physical image sets, with three executions per group
under the declared source/runtime/CPU profile. The flow comparison covers
three development families and one pretrained RAFT checkpoint. These designs do not
support population-level generalization or clinical performance claims.
The mechanical consumer demonstrates interpolation with supplied stiffness
and material identities, including deterioration under model mismatch; it
does not validate estimation of those inputs or recovery from ordinary images.
The image-only SfM comparisons use no FE prior. Binary agreement within
a fixed profile is not a test of agreement across rebuilds, hardware or thread
settings.

Resource users should state which quantities are provided to inference:
images and calibration, camera trajectories, surface geometry or mechanical
states. Supplying ground-truth poses or geometry changes the task.
Accuracy should be reported together with registration coverage, surface
support and missing outputs. Frames, points and overlapping windows do not
replace scene families as independent sampling units, and exposed development
data cannot also serve as unseen validation. Training on the three mechanical
bases and one material-parameter pair would not establish generalization to
other loads or tissues.
The proposed family allocation in Section~2.1 preserves these dependencies;
it does not restore an unseen status to families already inspected here.

The demonstrated scope is controlled tubular geometry relevant to colonoscopy.
Patient-specific anatomy, measured tissue properties, clinical optics,
large-deformation material validation, contact, fluids and lesions remain
outside that scope. Appearance realism, FE annotations and successful phantom
registration do not remove these limitations.

\section{Conclusion}
The Synthetic Deformable Tubular Benchmark provides linked image observations,
camera and surface states, persistent material correspondences and discrete
mechanical annotations. Paired acquisition and appearance conditions,
documented information boundaries and separately implemented consistency
checks support controlled evaluation against the generating models. The usage
studies show why reconstruction accuracy, registration coverage, surface
support and flow must be evaluated separately, and how mechanical records
can expose both benefits and mismatch of a prior. Fixed-environment
repeatability and retained-prediction metric replay remain distinct from
portable end-to-end reproduction.
By releasing both a fixed, validated benchmark and its open-source generator,
this work provides a reusable foundation for evaluating and extending
deformable reconstruction methods in controlled tubular environments.

\section{Data availability}
\label{sec:data-availability}
The Synthetic Deformable Tubular Benchmark, version \DatasetVersion{},
is publicly available \citep{cirauqui2026tubular} at \url{\DatasetDOI}.
The dataset distribution requires about 30.69 GB of downloads, including the
split archive and companion files. It is independent of the reproduction
material (about 8.75 GB of main files plus 1.13 GB of optional offline
dependencies); GB here denotes $10^9$ bytes.
Its baseline and appearance
checksum manifests identify the release. The accompanying data card,
schemas and validation reports describe the image, geometry and mechanical
records. Original generated data are licensed under Creative Commons
Attribution 4.0 International (CC BY 4.0); software and third-party works
retain separate terms.

Protocols, runners, pinned configurations, complete reconstructed models and
feature databases, flow predictions, evaluation records and provenance are
supplied in the reproduction package, version 1.1.0~\citep{cirauqui2026reproducibility}.
Direct access: \url{\ReproductionPackageURL}.
The scripts regenerate derived feature databases and reconstructions in
their specified environments. Per-study source/runtime receipts distinguish
the supplied dataset baseline models from the corrected comparison. Additional
diagnostic/control records are separate from the reported populations and
do not replace the complete current outputs. The selected distribution preserves
complete final inputs and outputs, with compact records of earlier failed or
invalid attempts; superseded raw-stage databases and build intermediates are
excluded. Reevaluation of the final outputs is therefore distinct from a
byte-for-byte reenactment of every historical stage.

\section{Code availability}
\label{sec:code-availability}
\texttt{LumenMorph} \LumenMorphVersion\ provides the generator, dataset reader,
prediction evaluator, usage-study runners, regression tests and source
verification tools at \url{\LumenMorphURL}. Its mechanics dependency,
\texttt{ContinuumFEM}, is available at \url{\ContinuumFEMURL} and pinned as a
Git submodule; clone with \texttt{--recurse-submodules}. Both use the MIT
license. Source snapshots and per-study dependency pins accompany
reproduction records.
Historical producer snapshots retain their own source hashes and remain
distinct from the current software release.

Pipeline A derives from base revision \texttt{\ExperimentalCOLMAPRevision};
its corrected source and native tests are available in the fixed release at
\url{\ExperimentalCOLMAPURL}. Its upstream colonoscopy adaptation is cited separately~\citep{subiascolmap,
subias2021thesis}; Pipeline A includes subsequent thesis developments rather
than being identical to that upstream repository. Pipeline B uses a corrected
derivative of upstream COLMAP \OfficialCOLMAPVersion\
(\url{https://github.com/colmap/colmap}); its public source and native tests are
fixed at \url{\OfficialCOLMAPForkURL}. The fork documents its execution-order
corrections and retains the original attribution and license.
The reproduction package records build flags, dependencies, source revisions
and executable hashes for both. Dataset-generation and corrected-comparison
profiles are identified separately; retained-prediction score replay and
fresh end-to-end execution are
distinguished in the reproduction records.
Supplementary Methods S1--S2 describe the reconstruction corrections,
CPU-assignment contract and lossless pose-export qualification. Complete
per-profile patches, binaries, native tests and positive/negative controls
remain in the reproduction package; conversion utilities do not replace the
frozen inference binaries.
The manuscript source archive includes the figure assets required to compile
the article.

\section{Declarations}
\noindent\textbf{Author contributions.} \AuthorContributionsStatement\par\smallskip
\textbf{Ethics.} \EthicsStatement\par\smallskip
\textbf{Funding.} \FundingStatement\par\smallskip
\textbf{Competing interests.} \CompetingInterestsStatement

\section*{Declaration of generative AI and AI-assisted technologies in the manuscript preparation process}
\AIDeclarationStatement

\begingroup
\let\OriginalDatasetBibitem\bibitem
\renewcommand{\bibitem}[2][]{%
  \OriginalDatasetBibitem[#1]{#2}%
  \def\CurrentDatasetKey{#2}%
  \def\PrimaryDatasetKey{cirauqui2026tubular}%
  \def\ReproductionDatasetKey{cirauqui2026reproducibility}%
  \ifx\CurrentDatasetKey\PrimaryDatasetKey [dataset]\space\fi
  \ifx\CurrentDatasetKey\ReproductionDatasetKey [dataset]\space\fi
}

\small
\raggedright
\setlength{\bibsep}{3pt}

\endgroup

\begin{thebibliography}{33}
\expandafter\ifx\csname natexlab\endcsname\relax\def\natexlab#1{#1}\fi
\providecommand{\url}[1]{\texttt{#1}}
\providecommand{\href}[2]{#2}
\providecommand{\path}[1]{#1}
\providecommand{\DOIprefix}{doi:}
\providecommand{\ArXivprefix}{arXiv:}
\providecommand{\URLprefix}{URL: }
\providecommand{\Pubmedprefix}{pmid:}
\providecommand{\doi}[1]{\href{http://dx.doi.org/#1}{\path{#1}}}
\providecommand{\Pubmed}[1]{\href{pmid:#1}{\path{#1}}}
\providecommand{\bibinfo}[2]{#2}
\ifx\xfnm\relax \def\xfnm[#1]{\unskip,\space#1}\fi
%Type = Misc
\bibitem[{Bleyer(n.d.)}]{bleyervoigt}
\bibinfo{author}{Bleyer, J.}, \bibinfo{year}{n.d.}
\newblock \bibinfo{title}{Isotropic and orthotropic plane stress elasticity}.
\newblock \bibinfo{howpublished}{Computational Mechanics Numerical Tours with
  FEniCSx}.
\newblock \URLprefix
  \url{https://bleyerj.github.io/comet-fenicsx/tours/linear_problems/isotropic_orthotropic_elasticity/isotropic_orthotropic_elasticity.html}.
  \bibinfo{note}{tensor and engineering Voigt conventions; accessed 12 October
  2026}.
%Type = Article
\bibitem[{Bobrow et~al.(2023)Bobrow, Golhar, Vijayan, Akshintala, Garcia and
  Durr}]{bobrow2023c3vd}
\bibinfo{author}{Bobrow, T.L.}, \bibinfo{author}{Golhar, M.},
  \bibinfo{author}{Vijayan, R.}, \bibinfo{author}{Akshintala, V.S.},
  \bibinfo{author}{Garcia, J.R.}, \bibinfo{author}{Durr, N.J.},
  \bibinfo{year}{2023}.
\newblock \bibinfo{title}{Colonoscopy {3D} video dataset with paired depth from
  {2D-3D} registration}.
\newblock \bibinfo{journal}{Med. Image Anal.} \bibinfo{volume}{90},
  \bibinfo{pages}{102956}.
\newblock \URLprefix \url{https://durrlab.github.io/C3VD/},
  \DOIprefix\doi{10.1016/j.media.2023.102956}.
%Type = Masterthesis
\bibitem[{Cirauqui~Viloria(2022)}]{cirauqui2022thesis}
\bibinfo{author}{Cirauqui~Viloria, {\'{I}}.}, \bibinfo{year}{2022}.
\newblock \bibinfo{title}{Deformable {Structure from Motion} in medical
  endoscopy sequences}.
\newblock \bibinfo{type}{Trabajo fin de m{\'a}ster}. Universidad de Zaragoza,
  Escuela de Ingenier{\'i}a y Arquitectura.
\newblock \bibinfo{note}{Spanish title: Structure from Motion deformable en
  secuencias de endoscopia m{\'e}dica; M{\'a}ster Universitario en
  Ingenier{\'i}a Industrial}.
%Type = Misc
\bibitem[{Cirauqui-Viloria(2026a)}]{lumenmorph2026}
\bibinfo{author}{Cirauqui-Viloria, {\'I}.}, \bibinfo{year}{2026}a.
\newblock \bibinfo{title}{{LumenMorph}: synthetic tubular data generation and
  evaluation}.
\newblock \bibinfo{howpublished}{Software, version 1.0.2}.
\newblock \URLprefix \url{https://github.com/icirauqui/LumenMorph/tree/v1.0.2}.
  \bibinfo{note}{accessed 10 October 2026}.
%Type = Misc
\bibitem[{Cirauqui-Viloria(2026b)}]{cirauqui2026reproducibility}
\bibinfo{author}{Cirauqui-Viloria, {\'I}.}, \bibinfo{year}{2026}b.
\newblock \bibinfo{title}{{Synthetic Deformable Tubular Benchmark:
  Reproducibility Material}}.
\newblock \bibinfo{howpublished}{Hugging Face, reproducibility material,
  version 1.1.0}.
\newblock \DOIprefix\doi{10.57967/hf/10851}.
%Type = Misc
\bibitem[{Cirauqui-Viloria(2026c)}]{cirauqui2026tubular}
\bibinfo{author}{Cirauqui-Viloria, {\'I}.}, \bibinfo{year}{2026}c.
\newblock \bibinfo{title}{{Synthetic Deformable Tubular Benchmark with Dense
  Ground Truth and Mechanical Annotations}}.
\newblock \bibinfo{howpublished}{Hugging Face, dataset, version 1.0.0}.
\newblock \DOIprefix\doi{10.57967/hf/10850}.
%Type = Misc
\bibitem[{{Durr Lab}(n.d.)}]{c3vddefcolrelease}
\bibinfo{author}{{Durr Lab}}, \bibinfo{year}{n.d.}
\newblock \bibinfo{title}{{C3VD-DEFCOL}: dataset release repository}.
\newblock \URLprefix \url{https://github.com/DurrLab/C3VDv3}.
  \bibinfo{note}{accessed 6 October 2026}.
%Type = Misc
\bibitem[{{DurrLab}(n.d.)}]{c3vdsource}
\bibinfo{author}{{DurrLab}}, \bibinfo{year}{n.d.}
\newblock \bibinfo{title}{{C3VD}: source code and example data loader}.
\newblock \URLprefix \url{https://github.com/DurrLab/C3VD}.
  \bibinfo{note}{exact source snapshot retained in the reproduction records;
  accessed 21 October 2026}.
%Type = Inproceedings
\bibitem[{Farneb{\"a}ck(2003)}]{farneback2003polynomial}
\bibinfo{author}{Farneb{\"a}ck, G.}, \bibinfo{year}{2003}.
\newblock \bibinfo{title}{Two-frame motion estimation based on polynomial
  expansion}, in: \bibinfo{booktitle}{Image Analysis (SCIA 2003)},
  \bibinfo{publisher}{Springer}. pp. \bibinfo{pages}{363--370}.
\newblock \URLprefix
  \url{https://link.springer.com/chapter/10.1007/3-540-45103-X_50},
  \DOIprefix\doi{10.1007/3-540-45103-X_50}.
%Type = Article
\bibitem[{Golhar et~al.(2026)Golhar, Galeano~Fretes, Ayers, Akshintala, Bobrow
  and Durr}]{golhar2026c3vdv2}
\bibinfo{author}{Golhar, M.V.}, \bibinfo{author}{Galeano~Fretes, L.S.},
  \bibinfo{author}{Ayers, L.}, \bibinfo{author}{Akshintala, V.S.},
  \bibinfo{author}{Bobrow, T.L.}, \bibinfo{author}{Durr, N.J.},
  \bibinfo{year}{2026}.
\newblock \bibinfo{title}{{C3VDv2}---colonoscopy {3D} video dataset with
  enhanced realism}.
\newblock \bibinfo{journal}{Sci. Data} \bibinfo{volume}{13},
  \bibinfo{pages}{1273}.
\newblock \URLprefix \url{https://www.nature.com/articles/s41597-026-07471-1},
  \DOIprefix\doi{10.1038/s41597-026-07471-1}.
%Type = Article
\bibitem[{Hong et~al.(2014)Hong, Tavanapong, Wong, Oh and
  de~Groen}]{hong2014virtualcolon}
\bibinfo{author}{Hong, D.}, \bibinfo{author}{Tavanapong, W.},
  \bibinfo{author}{Wong, J.}, \bibinfo{author}{Oh, J.H.},
  \bibinfo{author}{de~Groen, P.C.}, \bibinfo{year}{2014}.
\newblock \bibinfo{title}{{3D} reconstruction of virtual colon structures from
  colonoscopy images}.
\newblock \bibinfo{journal}{Comput. Med. Imaging Graph.} \bibinfo{volume}{38},
  \bibinfo{pages}{22--33}.
\newblock \URLprefix
  \url{https://www.sciencedirect.com/science/article/pii/S0895611113001523},
  \DOIprefix\doi{10.1016/j.compmedimag.2013.10.005}.
%Type = Article
\bibitem[{Horn(1987)}]{horn1987orientation}
\bibinfo{author}{Horn, B.K.P.}, \bibinfo{year}{1987}.
\newblock \bibinfo{title}{Closed-form solution of absolute orientation using
  unit quaternions}.
\newblock \bibinfo{journal}{J. Opt. Soc. Am. A} \bibinfo{volume}{4},
  \bibinfo{pages}{629--642}.
\newblock \URLprefix
  \url{https://opg.optica.org/josaa/abstract.cfm?URI=josaa-4-4-629},
  \DOIprefix\doi{10.1364/JOSAA.4.000629}.
%Type = Article
\bibitem[{Incetan et~al.(2021)Incetan, Celik, Obeid, Gokceler, Ozyoruk,
  Almalioglu, Chen, Mahmood, Gilbert, Durr and Turan}]{incetan2021vrcaps}
\bibinfo{author}{Incetan, K.}, \bibinfo{author}{Celik, I.O.},
  \bibinfo{author}{Obeid, A.}, \bibinfo{author}{Gokceler, G.I.},
  \bibinfo{author}{Ozyoruk, K.B.}, \bibinfo{author}{Almalioglu, Y.},
  \bibinfo{author}{Chen, R.J.}, \bibinfo{author}{Mahmood, F.},
  \bibinfo{author}{Gilbert, H.}, \bibinfo{author}{Durr, N.J.},
  \bibinfo{author}{Turan, M.}, \bibinfo{year}{2021}.
\newblock \bibinfo{title}{{VR-Caps}: A virtual environment for capsule
  endoscopy}.
\newblock \bibinfo{journal}{Med. Image Anal.} \bibinfo{volume}{70},
  \bibinfo{pages}{101990}.
\newblock \URLprefix \url{https://durr.jhu.edu/pubs/2021-vr-caps/},
  \DOIprefix\doi{10.1016/j.media.2021.101990}.
%Type = Misc
\bibitem[{Jakob et~al.(2022a)Jakob, Speierer, Roussel, Nimier-David, Vicini,
  Zeltner, Nicolet, Crespo, Leroy and Zhang}]{mitsuba3renderer}
\bibinfo{author}{Jakob, W.}, \bibinfo{author}{Speierer, S.},
  \bibinfo{author}{Roussel, N.}, \bibinfo{author}{Nimier-David, M.},
  \bibinfo{author}{Vicini, D.}, \bibinfo{author}{Zeltner, T.},
  \bibinfo{author}{Nicolet, B.}, \bibinfo{author}{Crespo, M.},
  \bibinfo{author}{Leroy, V.}, \bibinfo{author}{Zhang, Z.},
  \bibinfo{year}{2022}a.
\newblock \bibinfo{title}{{Mitsuba 3 renderer}}.
\newblock \bibinfo{howpublished}{Software, version 3.9.1}.
\newblock \URLprefix \url{https://github.com/mitsuba-renderer/mitsuba3}.
  \bibinfo{note}{author-supplied citation; accessed 12 October 2026}.
%Type = Article
\bibitem[{Jakob et~al.(2022b)Jakob, Speierer, Roussel and
  Vicini}]{jakob2022drjit}
\bibinfo{author}{Jakob, W.}, \bibinfo{author}{Speierer, S.},
  \bibinfo{author}{Roussel, N.}, \bibinfo{author}{Vicini, D.},
  \bibinfo{year}{2022}b.
\newblock \bibinfo{title}{{Dr.Jit}: A just-in-time compiler for differentiable
  rendering}.
\newblock \bibinfo{journal}{ACM Trans. Graph.} \bibinfo{volume}{41}.
\newblock \URLprefix \url{https://rgl.epfl.ch/publications/Jakob2022DrJit},
  \DOIprefix\doi{10.1145/3528223.3530099}.
%Type = Inproceedings
\bibitem[{Kroeger et~al.(2016)Kroeger, Timofte, Dai and
  Van~Gool}]{kroeger2016dis}
\bibinfo{author}{Kroeger, T.}, \bibinfo{author}{Timofte, R.},
  \bibinfo{author}{Dai, D.}, \bibinfo{author}{Van~Gool, L.},
  \bibinfo{year}{2016}.
\newblock \bibinfo{title}{Fast optical flow using dense inverse search}, in:
  \bibinfo{booktitle}{Computer Vision -- ECCV 2016},
  \bibinfo{publisher}{Springer}. pp. \bibinfo{pages}{471--488}.
\newblock \URLprefix
  \url{https://link.springer.com/chapter/10.1007/978-3-319-46493-0_29},
  \DOIprefix\doi{10.1007/978-3-319-46493-0_29}.
%Type = Article
\bibitem[{Lena et~al.(2026)Lena, Milesi, Casella, Carlini, Norton, Martin,
  Scaglioni, Obstein, De~Sire, Spadaccini, Hassan, Valdastri and
  De~Momi}]{lena2026realsyncol}
\bibinfo{author}{Lena, C.}, \bibinfo{author}{Milesi, D.},
  \bibinfo{author}{Casella, A.}, \bibinfo{author}{Carlini, L.},
  \bibinfo{author}{Norton, J.C.}, \bibinfo{author}{Martin, J.},
  \bibinfo{author}{Scaglioni, B.}, \bibinfo{author}{Obstein, K.L.},
  \bibinfo{author}{De~Sire, R.}, \bibinfo{author}{Spadaccini, M.},
  \bibinfo{author}{Hassan, C.}, \bibinfo{author}{Valdastri, P.},
  \bibinfo{author}{De~Momi, E.}, \bibinfo{year}{2026}.
\newblock \bibinfo{title}{A high-fidelity synthetic colon dataset for {3D}
  reconstruction applications}.
\newblock \bibinfo{journal}{Sci. Data} \URLprefix
  \url{https://www.nature.com/articles/s41597-026-07832-w},
  \DOIprefix\doi{10.1038/s41597-026-07832-w}.
%Type = Article
\bibitem[{Liu et~al.(2026)Liu, Yu, Xiao, He and Liu}]{liu2026scale}
\bibinfo{author}{Liu, Y.}, \bibinfo{author}{Yu, D.}, \bibinfo{author}{Xiao,
  Y.}, \bibinfo{author}{He, T.}, \bibinfo{author}{Liu, L.},
  \bibinfo{year}{2026}.
\newblock \bibinfo{title}{Scale-consistent {3D} reconstruction in monocular
  colonoscopy via camera-intrinsics-guided learning}.
\newblock \bibinfo{journal}{Comput. Med. Imaging Graph.} \bibinfo{volume}{129},
  \bibinfo{pages}{102734}.
\newblock \URLprefix
  \url{https://www.sciencedirect.com/science/article/pii/S0895611126000376},
  \DOIprefix\doi{10.1016/j.compmedimag.2026.102734}.
%Type = Misc
\bibitem[{Luk et~al.(2026)Luk, Golhar, Song, Iranzo, Batlle, Seenivasan,
  Montiel and Durr}]{luk2026c3vddefcol}
\bibinfo{author}{Luk, E.}, \bibinfo{author}{Golhar, M.V.},
  \bibinfo{author}{Song, A.}, \bibinfo{author}{Iranzo, R.},
  \bibinfo{author}{Batlle, V.M.}, \bibinfo{author}{Seenivasan, L.},
  \bibinfo{author}{Montiel, J.M.M.}, \bibinfo{author}{Durr, N.J.},
  \bibinfo{year}{2026}.
\newblock \bibinfo{title}{{C3VD-DEFCOL}: A deformable colonoscopy dataset with
  time-resolved {3D} ground truth and realistic appearance}.
\newblock \URLprefix \url{https://arxiv.org/abs/2606.07891},
  \DOIprefix\doi{10.48550/arXiv.2606.07891},
  \href{http://arxiv.org/abs/2606.07891}{{\tt arXiv:2606.07891}}.
  \bibinfo{note}{preprint; version 1}.
%Type = Inproceedings
\bibitem[{Martyniak et~al.(2025)Martyniak, Kaleta, Dall'Alba, Naskr{\k{e}}t,
  P{\l}otka and Korzeniowski}]{martyniak2025simuscope}
\bibinfo{author}{Martyniak, S.}, \bibinfo{author}{Kaleta, J.},
  \bibinfo{author}{Dall'Alba, D.}, \bibinfo{author}{Naskr{\k{e}}t, M.},
  \bibinfo{author}{P{\l}otka, S.}, \bibinfo{author}{Korzeniowski, P.},
  \bibinfo{year}{2025}.
\newblock \bibinfo{title}{{SimuScope}: Realistic endoscopic synthetic dataset
  generation through surgical simulation and diffusion models}, in:
  \bibinfo{booktitle}{Proceedings of the Winter Conference on Applications of
  Computer Vision (WACV)}, pp. \bibinfo{pages}{4268--4278}.
\newblock \URLprefix
  \url{https://openaccess.thecvf.com/content/WACV2025/papers/Martyniak_SimuScope_Realistic_Endoscopic_Synthetic_Dataset_Generation_through_Surgical_Simulation_and_WACV_2025_paper.pdf}.
%Type = Misc
\bibitem[{{Mitsuba contributors}(2026)}]{mitsuba3bsdf}
\bibinfo{author}{{Mitsuba contributors}}, \bibinfo{year}{2026}.
\newblock \bibinfo{title}{{Mitsuba 3}: Bidirectional scattering distribution
  functions}.
\newblock \URLprefix
  \url{https://mitsuba.readthedocs.io/en/stable/src/generated/plugins_bsdfs.html}.
  \bibinfo{note}{official documentation; roughplastic and diffuse plugins;
  accessed 12 September 2026}.
%Type = Misc
\bibitem[{{PyTorch contributors}(2026)}]{torchvisionraft}
\bibinfo{author}{{PyTorch contributors}}, \bibinfo{year}{2026}.
\newblock \bibinfo{title}{{torchvision.models.optical\_flow}: {RAFT} reference
  implementation and pretrained weights}.
\newblock \URLprefix
  \url{https://github.com/pytorch/vision/tree/main/references/optical_flow}.
  \bibinfo{note}{\texttt{raft\_large}, weights
  \texttt{Raft\_Large\_Weights.C\_T\_SKHT\_V2}; accessed 15 September 2026}.
%Type = Article
\bibitem[{Rau et~al.(2024)Rau, Bano, Jin, Azagra, Morlana, Kader, Sanderson,
  Matuszewski, Lee, Lee, Posner, Frank, Elangovan, Raviteja, Li, Liu,
  Lalithkumar, Islam, Ren, Lovat, Montiel and Stoyanov}]{rau2024simcol3d}
\bibinfo{author}{Rau, A.}, \bibinfo{author}{Bano, S.}, \bibinfo{author}{Jin,
  Y.}, \bibinfo{author}{Azagra, P.}, \bibinfo{author}{Morlana, J.},
  \bibinfo{author}{Kader, R.}, \bibinfo{author}{Sanderson, E.},
  \bibinfo{author}{Matuszewski, B.J.}, \bibinfo{author}{Lee, J.Y.},
  \bibinfo{author}{Lee, D.J.}, \bibinfo{author}{Posner, E.},
  \bibinfo{author}{Frank, N.}, \bibinfo{author}{Elangovan, V.},
  \bibinfo{author}{Raviteja, S.}, \bibinfo{author}{Li, Z.},
  \bibinfo{author}{Liu, J.}, \bibinfo{author}{Lalithkumar, S.},
  \bibinfo{author}{Islam, M.}, \bibinfo{author}{Ren, H.},
  \bibinfo{author}{Lovat, L.B.}, \bibinfo{author}{Montiel, J.M.M.},
  \bibinfo{author}{Stoyanov, D.}, \bibinfo{year}{2024}.
\newblock \bibinfo{title}{{SimCol3D}---{3D} reconstruction during colonoscopy
  challenge}.
\newblock \bibinfo{journal}{Med. Image Anal.} \bibinfo{volume}{96},
  \bibinfo{pages}{103195}.
\newblock \URLprefix \url{https://discovery.ucl.ac.uk/id/eprint/10194104/},
  \DOIprefix\doi{10.1016/j.media.2024.103195}.
%Type = Article
\bibitem[{Ruano et~al.(2024)Ruano, G{\'o}mez, Romero and
  Manzanera}]{ruano2024sfs}
\bibinfo{author}{Ruano, J.}, \bibinfo{author}{G{\'o}mez, M.},
  \bibinfo{author}{Romero, E.}, \bibinfo{author}{Manzanera, A.},
  \bibinfo{year}{2024}.
\newblock \bibinfo{title}{Leveraging a realistic synthetic database to learn
  {Shape-from-Shading} for estimating the colon depth in colonoscopy images}.
\newblock \bibinfo{journal}{Comput. Med. Imaging Graph.} \bibinfo{volume}{115},
  \bibinfo{pages}{102390}.
\newblock \URLprefix
  \url{https://www.sciencedirect.com/science/article/pii/S0895611124000673},
  \DOIprefix\doi{10.1016/j.compmedimag.2024.102390}.
%Type = Inproceedings
\bibitem[{Sch{\"o}nberger and Frahm(2016)}]{schonberger2016colmap}
\bibinfo{author}{Sch{\"o}nberger, J.L.}, \bibinfo{author}{Frahm, J.M.},
  \bibinfo{year}{2016}.
\newblock \bibinfo{title}{Structure-from-motion revisited}, in:
  \bibinfo{booktitle}{Proceedings of the IEEE Conference on Computer Vision and
  Pattern Recognition (CVPR)}, pp. \bibinfo{pages}{4104--4113}.
\newblock \URLprefix
  \url{https://openaccess.thecvf.com/content_cvpr_2016/html/Schonberger_Structure-From-Motion_Revisited_CVPR_2016_paper.html},
  \DOIprefix\doi{10.1109/CVPR.2016.445}.
%Type = Misc
\bibitem[{{SciPy contributors}(2026)}]{scipyrbf}
\bibinfo{author}{{SciPy contributors}}, \bibinfo{year}{2026}.
\newblock \bibinfo{title}{{RBFInterpolator}: radial-basis interpolation
  documentation}.
\newblock \URLprefix
  \url{https://docs.scipy.org/doc/scipy/reference/generated/scipy.interpolate.RBFInterpolator.html}.
  \bibinfo{note}{accessed 6 October 2026; implementation used: SciPy 1.18.1}.
%Type = Misc
\bibitem[{{SciPy contributors}(n.d.)}]{scipycsr}
\bibinfo{author}{{SciPy contributors}}, \bibinfo{year}{n.d.}
\newblock \bibinfo{title}{{csr\_matrix}: compressed sparse row matrix
  representation}.
\newblock \URLprefix
  \url{https://docs.scipy.org/doc/scipy/reference/generated/scipy.sparse.csr_matrix.html}.
  \bibinfo{note}{official documentation; accessed 12 October 2026}.
%Type = Article
\bibitem[{Singh et~al.(2025)Singh, Bhaumik and
  Chatterjee}]{singh2025simintestine}
\bibinfo{author}{Singh, S.}, \bibinfo{author}{Bhaumik, B.},
  \bibinfo{author}{Chatterjee, S.}, \bibinfo{year}{2025}.
\newblock \bibinfo{title}{{SimIntestine}: A synthetic dataset from virtual
  capsule endoscope}.
\newblock \bibinfo{journal}{Med. Image Anal.} \bibinfo{volume}{105},
  \bibinfo{pages}{103706}.
\newblock \URLprefix
  \url{https://www.sciencedirect.com/science/article/pii/S1361841525002531},
  \DOIprefix\doi{10.1016/j.media.2025.103706}.
%Type = Misc
\bibitem[{Smolak-Dy{\.z}ewska et~al.(2026)Smolak-Dy{\.z}ewska, Kaleta,
  Dall'Alba and Spurek}]{smolak2026colonsplat}
\bibinfo{author}{Smolak-Dy{\.z}ewska, W.}, \bibinfo{author}{Kaleta, J.},
  \bibinfo{author}{Dall'Alba, D.}, \bibinfo{author}{Spurek, P.},
  \bibinfo{year}{2026}.
\newblock \bibinfo{title}{{ColonSplat}: Reconstruction of peristaltic motion in
  colonoscopy with dynamic gaussian splatting}.
\newblock \URLprefix \url{https://arxiv.org/abs/2603.06860},
  \DOIprefix\doi{10.48550/arXiv.2603.06860},
  \href{http://arxiv.org/abs/2603.06860}{{\tt arXiv:2603.06860}}.
  \bibinfo{note}{preprint; version 1}.
%Type = Masterthesis
\bibitem[{Sub{\'i}as~Sarrato(2021)}]{subias2021thesis}
\bibinfo{author}{Sub{\'i}as~Sarrato, J.D.}, \bibinfo{year}{2021}.
\newblock \bibinfo{title}{Modelado {3D} del colon a partir de colonoscopias
  reales mediante geometr{\'i}a multivista}.
\newblock \bibinfo{type}{Trabajo fin de grado}. Universidad de Zaragoza,
  Escuela de Ingenier{\'i}a y Arquitectura.
\newblock \URLprefix
  \url{https://zaguan.unizar.es/record/106905/files/TAZ-TFG-2021-2094.pdf}.
%Type = Misc
\bibitem[{Sub{\'i}as~Sarrato(n.d.)}]{subiascolmap}
\bibinfo{author}{Sub{\'i}as~Sarrato, J.D.}, \bibinfo{year}{n.d.}
\newblock \bibinfo{title}{{COLMAP} for colonoscopy}.
\newblock \bibinfo{howpublished}{UZ-SLAMLab software repository}.
\newblock \URLprefix
  \url{https://github.com/UZ-SLAMLab/colmap_for_colonoscopy}.
  \bibinfo{note}{accessed 30 September 2026}.
%Type = Inproceedings
\bibitem[{Teed and Deng(2020)}]{teed2020raft}
\bibinfo{author}{Teed, Z.}, \bibinfo{author}{Deng, J.}, \bibinfo{year}{2020}.
\newblock \bibinfo{title}{{RAFT}: Recurrent all-pairs field transforms for
  optical flow}, in: \bibinfo{booktitle}{Computer Vision -- ECCV 2020},
  \bibinfo{publisher}{Springer}. pp. \bibinfo{pages}{402--419}.
\newblock \URLprefix
  \url{https://link.springer.com/chapter/10.1007/978-3-030-58536-5_24},
  \DOIprefix\doi{10.1007/978-3-030-58536-5_24}.
%Type = Article
\bibitem[{Umeyama(1991)}]{umeyama1991similarity}
\bibinfo{author}{Umeyama, S.}, \bibinfo{year}{1991}.
\newblock \bibinfo{title}{Least-squares estimation of transformation parameters
  between two point patterns}.
\newblock \bibinfo{journal}{IEEE Trans. Pattern Anal. Mach. Intell.}
  \bibinfo{volume}{13}, \bibinfo{pages}{376--380}.
\newblock \URLprefix
  \url{https://web.stanford.edu/class/cs273/refs/umeyama.pdf},
  \DOIprefix\doi{10.1109/34.88573}.

\end{thebibliography}
\end{document}

\section*{Supplementary Methods}
\SubmissionIdentity
\section*{S1. Qualified reconstruction profiles and source repairs}
This supplement preserves implementation details behind the two pipelines
reported in the main article. Both pipelines receive identical public
images, calibration and requested windows; algorithmic causes cannot be
assigned to a single repair from this cross-pipeline comparison.

The pipelines share COLMAP ancestry and are not unrelated SfM methods.
A-RNG1, A-SIFT1, B-INIT1 and B-ORDER1 name cumulative source/execution
profiles, not four independent reconstruction algorithms. A-SIFT1 extends
A-RNG1 in Pipeline A; B-ORDER1 extends B-INIT1 in Pipeline B. The corrected
functionality is included in the two public repositories linked in
the main article (Code availability); the reproduction package supplies the
exact per-profile source patches, binaries and configuration receipts.
The summary below separates cumulative repairs from fresh and retained stages;
Supplementary Methods S2 details the exact-input reuse checks.
\begin{center}
\manuscripttablestyle
\begin{tabularx}{\linewidth}{@{}>{\raggedright\arraybackslash}p{0.15\linewidth}YY@{}}
\toprule
Pipeline / final profile & Cumulative repairs & Execution lineage \\
\midrule
A / A-SIFT1 &
A-RNG1: shared CLI seed and thread-local FLANN sampling/shuffling, avoiding
process-global seed resets. A-SIFT1: initializes unwritten CPU descriptor
buffers at unsupported image/octave boundaries; empty
RootSIFT rows remain finite zeros. &
Fresh extraction and mapping. Matching is reused only when canonical feature
inputs agree exactly; otherwise it is recomputed. \\
B / B-ORDER1 &
B-INIT1: declared initial-pair search/history order and image-ID tie breaking.
B-ORDER1: additionally sorts bundle-adjustment residuals by image and point
identity. &
Appearance: retained B-INIT1 extraction/matching, fresh B-ORDER1 mapping.
Physical: retained B-ORDER1 matching, fresh B-ORDER1 mapping. \\
\bottomrule
\end{tabularx}
\end{center}
Valid nonempty descriptor arithmetic and keypoint geometry are unchanged in A.
The B repairs preserve upstream geometry thresholds, the optimization objective
and model-size policy. Retained extraction/matching is not fresh end-to-end
reproduction; its profile-specific lineage accompanies the outputs.

Pipeline A inherits calibrated two-view geometry and guided-matching changes
from the colonoscopy adaptation, with later discarded-track diagnostics.
Its base revision and applied profile patches accompany the reproduction material.

\clearpage
\section*{S2. Runtime, exports and reproduction scope}
The separately qualified A-EXPORT1 correction increases text precision;
A-EXPORT2 adds lossless binary-pose conversion without quaternion
renormalization. These conversion tools do not replace the frozen inference
binaries. Source patches and qualification records accompany the reproduction package.

The export utility writes 17 significant digits with a fixed locale, and every
exported pose is checked for exact agreement with its full-precision binary
values. A separate audit reads every binary model, fits the proper positive
similarity by a quaternion eigensystem, evaluates rotations with SciPy and
computes local surface distances with a triangle bounding-box tree.
Manufactured positive and negative controls precede the full-cohort audit.
This validates the scoring calculation against the supplied reference,
not physical accuracy beyond it.

Fixed source/executable/dependency hashes, seed zero (including B's mapper-estimator seed), one requested native,
OpenMP and BLAS thread, disabled intrinsic refinement, effective mapper
options, private matched databases and the recorded CPU affinity define the
repeatability contract. CPU-AFFINITY1 specifies the Ubuntu resource assignment;
its homogeneous CPU/cache profile is part of that contract, not a general
hardware-portability guarantee. Numerical equality of three runs in one
profile does not establish equality across hardware, operating systems,
compilers or rebuilt dependencies.

Actual Docker CPU sets, native
arguments and cache-class readbacks are retained. An identical-operand
control demonstrates that Eigen's hardware-dependent matrix-product blocking
can change floating-point rounding between core classes; no inference
source, numerical precision, seed or threshold changes under this profile.

The retained source qualification includes native positive/negative controls
for the original numerical initialization and ordering defects and separate
patch application/readback. Failed, invalid and interrupted attempts remain
in compact provenance records. The final published inputs, models,
predictions, metrics and separate feature databases are retained; superseded
raw-stage databases and build intermediates are not. Recomputing scores from
retained predictions, rerunning a specified inference stage and reproducing
the complete historical workflow are different claims.

\subsection*{Stage identity and execution checks}
The repeatability contract requires three A-SIFT1 executions and three
corrected B global/local mapper repetitions for every primary family and
condition under the frozen one-thread, seed-zero profile. A extracts features
afresh and preserves each extraction database. Matching is inherited only when
all canonical camera, image, keypoint and descriptor schemas, identities, shapes
and payloads equal the corresponding closed matching input exactly; otherwise
it is recomputed. All A mappers are fresh. B inherits its corresponding closed
extraction/matching inputs. A separately implemented direct-row readback verifies every
A reuse decision and all private mapper databases. We compare canonical complete SQLite payloads, every
full-precision binary model file, stage exits, effective native options and actual
registered frame names. Documentary output paths are recorded separately.
This tests repeatability on the complete requested population; a fixed seed
or one successfully repeated example alone is insufficient. Timeouts,
fragments and absent models remain in the denominator. Deterministic data
generation, inference repeatability, coverage and metric replay are separate
properties. The three runs are computational repetitions, not independent
scene observations or a hardware-portability test. The current A-SIFT1 and
corrected B populations use a dedicated 24-core Intel Core Ultra 9 Ubuntu
host with 64 GB RAM. Independent physical groups may execute
concurrently, while the three repetitions within each group remain
sequential. The Ubuntu processor has two core/cache classes. The
\textbf{CPU-AFFINITY1} execution profile restricts every native process
in a three-repeat group to the same single core from the homogeneous
eight-core class, including its children. Runtime receipts retain the actual CPU sets and effective options.
The hardware-blocking control is described above.
The one-thread numerical profile is unchanged. Host differences
and resource contention preclude a comparative wall-clock speed claim.

\clearpage
\section*{S3. Extending the dataset with the published generator}
Use the fixed LumenMorph release and its pinned mechanics submodule, following
its numerical installation guide. Copy and edit the JSON specification before
freezing a new suite. The commands illustrate the published interface for f01 and its FEM condition;
other declared conditions use the same staged interface.
\begingroup\small
\begin{verbatim}
git clone --branch v1.0.2 --recurse-submodules \
  https://github.com/icirauqui/LumenMorph.git
cd LumenMorph
# Install the pinned numerical environment as documented in the release.
# Copy configs/tubular_benchmark_v1.json to custom.json and edit it.
python -m backend.src.benchmark freeze \
  --root ./new_suite --spec custom.json
python -m backend.src.benchmark prepare \
  --root ./new_suite --family f01
python -m backend.src.benchmark render \
  --root ./new_suite --family f01 --stream reference
python -m backend.src.benchmark package \
  --root ./new_suite --family f01 --condition fem
python -m backend.src.benchmark render \
  --root ./new_suite --family f01 --condition fem --stream stationary
python -m backend.src.benchmark render \
  --root ./new_suite --family f01 --condition fem --stream moving
\end{verbatim}
\endgroup
Use \texttt{python} from the installed numerical environment. Each stage
requires a fresh output location and verifies the frozen sources. New suites
have their own recorded specification, provenance and checksums. The
published benchmark producer uses a fixed 90-state motion schedule, even
when spatial configuration, load amplitudes and camera parameters are changed.
Geometry modifications must remain within the generator's supported tubular
parameterization and pass their own geometric and mechanical checks.

\paragraph{Output path for this example.}
The commands above generate the f01 FEM condition at \path{./new_suite/v4}.
The following consumption example instead opens the fixed release at
\texttt{benchmark/v2}; these numbers are not interchangeable.
\texttt{PublicDataset} and \texttt{EvaluationDataset} from
\path{backend.src.benchmark_reader} accept the generated path
\path{./new_suite/v4} for this example. The producer and
release use different numbering: the released f01 FEM package is
\texttt{benchmark/v2}. Use the generated directory containing
\texttt{DATASET.json}, rather than assuming release numbering. The former reader accesses the public images, calibration
and acquisition metadata; the latter additionally exposes the explicitly
privileged geometry, persistent material queries and applicable FE states.
The two appearance readers serve the separate matched-appearance schema.
These information boundaries must also be retained when generating a custom
suite. Newly generated scenes are not part of the fixed release or covered
automatically by its validation reports.
\clearpage
\section*{S4. Minimal consumption and prediction evaluation}
Use Python from the installed LumenMorph numerical environment, with its
repository root as the working directory so that \texttt{backend.src} imports
resolve. Download and reassemble the complete fixed dataset following its
release README. Set \texttt{SDTUBE\_DATA}
to the extracted directory containing \texttt{benchmark/}, and
\texttt{SDTUBE\_OUT} to a fresh writable directory outside it.
The package selected below is the released f01 FEM condition, not the
\texttt{v4} directory generated by S3.

The following complete example saves a zero-displacement control using only
public image dimensions, then opens privileged truth to query the same
material points at frames 52 and 53 and score the already saved prediction.
The baseline reader returns BGR channel order; only its shape is used here.
This control demonstrates the interface and is not a trained reconstruction
method or an additional scientific comparison.
\begingroup\small
\begin{verbatim}
import os
from pathlib import Path
import numpy as np
from backend.src.benchmark_reader import PublicDataset, EvaluationDataset
from backend.src.benchmark_evaluate import evaluate

package = Path(os.environ['SDTUBE_DATA']) / 'benchmark/v2'
output = Path(os.environ['SDTUBE_OUT'])
output.mkdir(parents=True, exist_ok=False)
public = PublicDataset(package)
# Freeze a zero-motion control using only the public image shape.
height, width = public.rgb(52, 'moving').shape[:2]
prediction = np.zeros((height, width, 3), dtype=np.float64)
np.savez_compressed(output / 'prediction.npz', prediction=prediction)

# Privileged inspection begins only after the prediction has been saved.
truth = EvaluationDataset(package)
dense = truth.dense(52, 'moving')
ids = dense['triangle_id'][dense['valid']][::4096]
weights = dense['material_barycentric'][dense['valid']][::4096]
x52 = truth.material_points(ids, weights, 52, 'moving')
x53 = truth.material_points(ids, weights, 53, 'moving')
print('Queried material points:', len(ids))
print('First point displacement:', x53[0] - x52[0])
scores = evaluate(package, output / 'prediction.npz', output / 'scores.json',
                  task='displacement', frame=52, target=53, stream='moving',
                  coordinate_frame='benchmark-world')
print(scores['primary'])
\end{verbatim}
\endgroup
The query uses immutable triangle IDs and barycentric weights from the source
frame, so \texttt{x53-x52} follows the same material, including points that may
be invisible later. The scorer evaluates every source-valid surface pixel;
the sparse inspection query does not select the scored population. The JSON
records coverage, conditional/full-population errors and input hashes. Its
world-frame declaration specifies axes and scene units; the scorer fits no
transform or scale to truth. A real method must save its own prediction before
privileged inspection, use only its declared inference inputs and retain
missing outputs in coverage. Prediction shape is native $H\times W\times3$;
other tasks and appearance-specific availability follow the main article.
\clearpage
\section*{S5. Common-support camera contrasts}
The main article reports camera accuracy alongside registration, surface
precision and support. This additional view retains every B-minus-A contrast
on common registered frames; it is not a separate benchmark or a controlled
ablation of a single algorithmic change.
\renewcommand{\thefigure}{S\arabic{figure}}
\setcounter{figure}{0}
\begin{figure}[htbp]
\centering
\begin{subfigure}[t]{0.49\linewidth}
\centering\includegraphics[width=\linewidth]{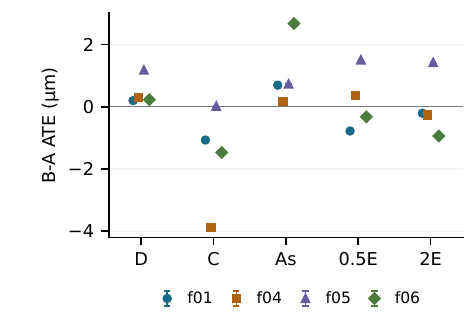}
\caption{Global.}
\end{subfigure}\hfill
\begin{subfigure}[t]{0.49\linewidth}
\centering\includegraphics[width=\linewidth]{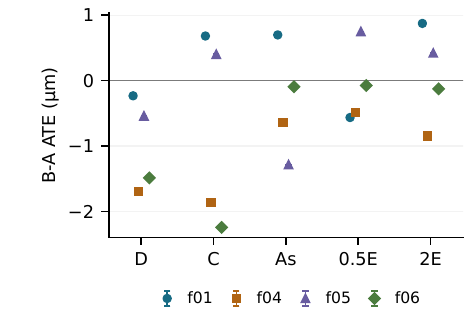}
\caption{Local.}
\end{subfigure}
\caption{B-minus-A camera-center ATE RMSE on common requested frames, in
micrometers. Positive values indicate larger error for B. Colors and marker
shapes identify families. D: diffuse symmetric; C: coated symmetric; As:
diffuse asymmetric; 0.5E and 2E: half and double exposure. Each model is fitted on the same support
for each contrast. Micrometer units express numerical differences on the
assigned synthetic scale after similarity fitting; they do not establish
micrometer physical measurement accuracy or absolute-scale recovery.
All three repeated contrasts coincide exactly.}
\label{fig:supplement-camera-difference}
\end{figure}